\documentclass[journal,twocolumn]{IEEEtran}

\usepackage{amsmath}
\usepackage{xcolor}
\usepackage{url}
\usepackage{epsfig}
\usepackage{multirow}
\usepackage{array}
\usepackage{cite}

\newcommand*{\BLUE}[1]{{\color{blue}#1}}
\newcommand*{\RED}[1]{{\color{red}#1}}
\definecolor{green1}{RGB}{0,176,80}
\definecolor{magenta1}{RGB}{255,0,255}
\newcommand*{\GREEN}[1]{{\textcolor{green1}{#1}}}
\newcommand*{\MG}[1]{{\textcolor{magenta1}{#1}}}
\newcommand*{\mb}[1]{{\mathbf{#1}}}

\def\R#1{(\ref{#1})}
\newcolumntype{P}[1]{>{\centering\arraybackslash}p{#1}}

\newenvironment{myquote}[1]%
  {\list{}{\leftmargin=#1\rightmargin=#1}\item[]}%
  {\endlist}

\usepackage{stackengine}
\def\deleq{\mathrel{\ensurestackMath{\stackon[1pt]{=}{\scriptstyle\Delta}}}}

\begin{document}

\title{
Improved Real-Time Monocular SLAM Using \\
Semantic Segmentation on Selective Frames}

\author{
Jinkyu Lee, \IEEEmembership{Student Member, IEEE,} Muhyun Back, \IEEEmembership{Student Member, IEEE,} 
\\
Sung Soo Hwang$^{\ast}$, \IEEEmembership{Member, IEEE,} and Il Yong Chun$^{\ast}$, \IEEEmembership{Member, IEEE}\vspace*{-0.5pc}

\thanks{S.~S.~Hwang is supported by Sabbatical Leave Grant from Handong Global University (HGU-2021).
I.~Y.~Chun is supported in part by 
the National Research Foundation of Korea (NRF) grant NRF-2022R1F1A1074546 funded by the Ministry of Science of ICT (MSIT),
the Institute of Information \& communications Technology Planning \& Evaluation (IITP) grant 2019-0-00421 funded by MSIT, 
the Institute for Basic Science grant (IBS) IBS-R015-D1,
the Technology Innovation Program grant 20014967 funded by the Ministry of Trade, Industry \& Energy (MOTIE),
and the Korea Institute for Advancement of Technology grant P0022094 funded by the Ministry of Education-MOTIE.}
\thanks{$^\ast$Corresponding authors}
\thanks{Jinkyu Lee and Muhyun Back are with the Department of Information and Communication Engineering, Handong Global University, Pohang 37554, South Korea (e-mails: 21931008@handong.edu; 21931005@handong.edu).}
\thanks{Sung Soo Hwang is with the School of Computer Science and Electrical Engineering, Handong Global University, Pohang 37554, South Korea (e-mail: sshwang@handong.edu).}
\thanks{I.~Y.~Chun is with the School of Electronic and Electrical Engineering and the Department of Artificial Intelligence, Sungkyunkwan University, and the Center for Neuroscience Imaging Research, IBS, Suwon, Gyeonggi-Do 16419, South Korea (e-mail: iychun@skku.edu).}
}

\maketitle

\begin{abstract}
Monocular simultaneous localization and mapping (SLAM) is emerging in advanced driver assistance systems and autonomous driving,
because a single camera is cheap and easy to install.
Conventional monocular SLAM has two major challenges leading inaccurate localization and mapping.
First, it is challenging to estimate scales in localization and mapping.
Second, conventional monocular SLAM uses inappropriate mapping factors such as dynamic objects and low-parallax areas in mapping.
This paper proposes an improved real-time monocular SLAM that resolves the aforementioned challenges by efficiently using deep learning-based semantic segmentation.
To achieve the real-time execution of the proposed method, we apply semantic segmentation only to downsampled keyframes in parallel with mapping processes.
In addition, the proposed method corrects scales of camera poses and three-dimensional (3D) points, using estimated ground plane from road-labeled 3D points and the real camera height.
The proposed method also removes inappropriate corner features labeled as moving objects and low parallax areas.
Experiments with eight video sequences demonstrate that 
the proposed monocular SLAM system achieves significantly improved and comparable trajectory tracking accuracy,
compared to existing state-of-the-art \emph{monocular} and \emph{stereo} SLAM systems, respectively.
The proposed system can achieve real-time tracking on a standard CPU potentially with a standard GPU support, 
whereas existing segmentation-aided monocular SLAM does not.
\end{abstract}


\section{Introduction}

\IEEEPARstart Simultaneous localization and mapping (SLAM) techniques have been evolving and widely applied to advanced driver assistance systems (ADAS) and autonomous driving systems.
While SLAM approaches using light detection and ranging (LiDAR) sensors are accurate, 
the cost of LiDAR sensors is high and they have not been widely used in commercial products.
Visual SLAM systems that use camera(s) are a popular alternative to LiDAR-based SLAM.
Monocular SLAM systems that use a single camera are attractive as they are cheap and easy to install. 
Monocular SLAM was initially suggested with filter-based approaches~\cite{davison2007monoslam,civera2008inverse,chiuso2002structure,eade2006scalable}.
The filter-based methods are computationally inefficient, since both localization and mapping run on every frame~\cite{younes2017keyframe}.
To resolve the issue of filter-based methods,
keyframe-based approaches \cite{mouragnon2006real,klein2007parallel,mur2015orb} (see other references in \cite{younes2017keyframe})
run the mapping process only on selective frames, called keyframes, 
while the localization process estimates a camera pose in every frame.
The keyframe-based SLAM improved the localization accuracy and computational efficiency of filter-based methods~\cite{strasdat2012visual}, 
and became the \textit{de facto} standard in monocular SLAM~\cite{younes2017keyframe}.

\begin{figure}
\centering
\includegraphics[width=1\linewidth]{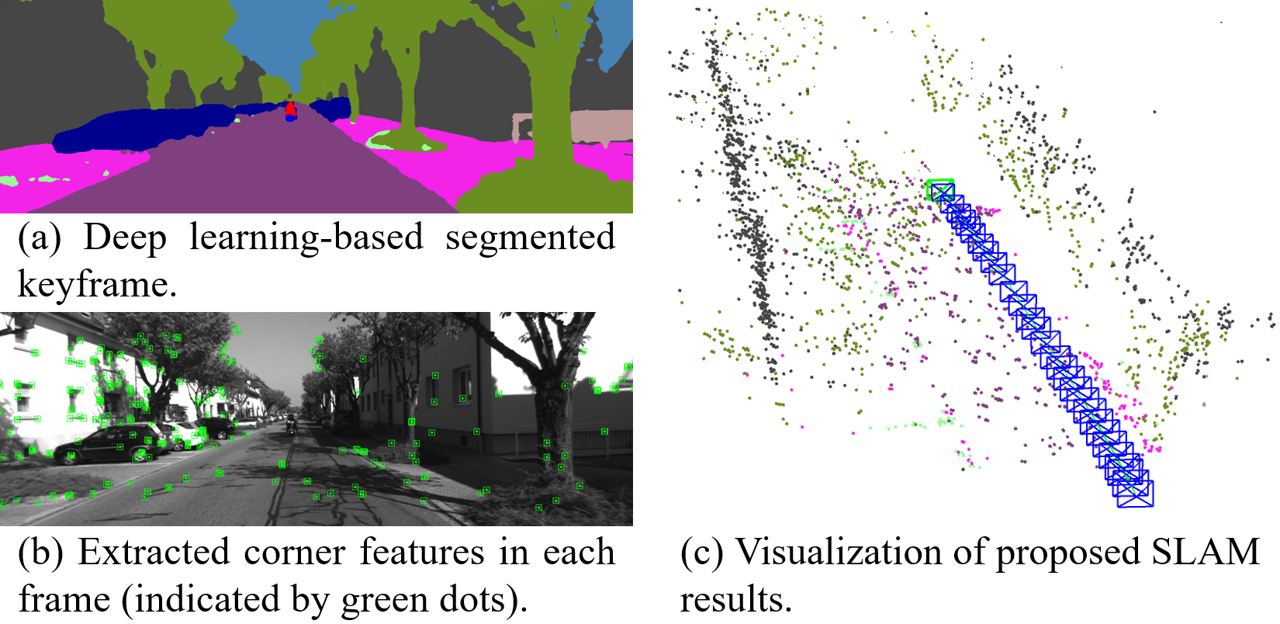}
\caption{
Intermediate results of the proposed monocular SLAM framework.
The proposed method uses segmented keyframes in (a), to choose appropriate corner features extracted in (b) and label reconstructed 3D points in (c).
In (c), the colors of labeled 3D points correspond to those in the segmentation result (b). 
The proposed method uses road-labeled 3D points (colored with purple in (c)) for scale estimation.
}
\label{fig_intermediate_result}
\end{figure}

\begin{figure*}
\centering
\includegraphics[width=0.85\linewidth]{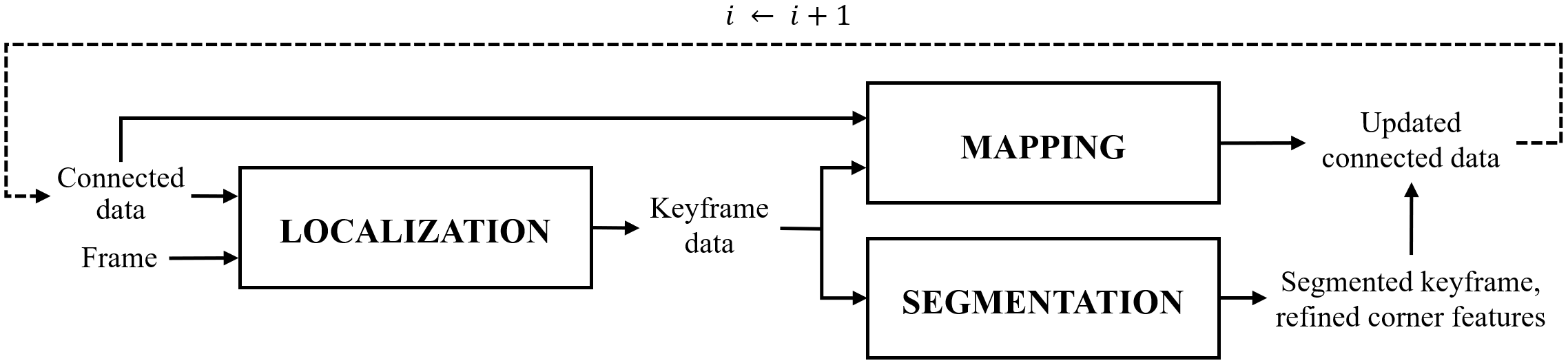}
\caption{
Overview of the proposed monocular SLAM system between previous and current keyframe time points, indexed by $i-1$ and $i$, respectively. 
The keyframe data consists of current keyframe and its corner features and camera pose, and previous keyframe and its corrected camera pose. 
These are denoted by 
$\BLUE{\mb{k}^{(i)}}$, $\RED{\mb{f}^{(i)}}$, $\GREEN{\mb{P}^{(i)}}$, $\BLUE{\mb{k}^{(i-1)}}$, $\GREEN{\hat{\mb{P}}^{(i-1)}}$,
respectively; see notations in Section~\ref{ss_preliminary}.
The connected data consists of refined corner features, corrected camera pose, segmentation result and corrected 3D points of connected keyframes.
These are denoted by 
$\RED{\hat{\mb{f}}^{(i')}}$, $\GREEN{\hat{\mb{P}}^{(i')}}$, $\BLUE{\tilde{\mb{k}}^{(i')}}$, $\MG{\hat{\mb{x}}^{(i')}}$ for $i' \in \mathcal{C}^{(i)}$, respectively; see notations in Section~\ref{ss_preliminary}.
The symbol colors correspond to those in Fig.~\ref{fig_method1}.
}
\label{fig_system_overview}
\end{figure*}

Conventional monocular SLAM has major challenges in scale estimation and mapping.
There are two scale estimation challenges in monocular SLAM, namely, scale ambiguity and scale drift~\cite{zhou2019ground}.
First, one cannot recover camera poses and three-dimensional (3D) points with absolute scale.
Second, errors are accumulated by following the SLAM procedures,
i.e., feature extraction, feature matching, camera pose estimation, and 3D structure reconstruction. 
In monocular SLAM, the scale ambiguity is the unique problem and scale drifts become more severe due to the scale ambiguity.
If the scale issues exist, monocular SLAM can mis-localize a current camera (or vehicle) position and generate 3D points in ``off-the-grid'' locations.
The other major challenge is inaccurate mapping from two-dimensional (2D) points to 3D points.
The mapping process of monocular SLAM consists of the triangulation of 2D point pairs obtained from two adjacent frames collected from different time points \cite{hartley2003multiple},
where we refer to a point in time as a time point.
For accurate mapping, static scenes and sufficient parallax are required.
In practice, however, some objects such as cars, pedestrians, etc. moves in a scene,
or some objects are distant from a camera with low-parallax.

\subsection{Related methods for overcoming limitations of monocular SLAM}
\label{sec:work}

This section reviews related works that attempted to resolve limitations of monocular SLAM.

To resolve the scale ambiguity issue, 
many existing methods use prior knowledge, e.g., camera height~\cite{zhou2019ground,grater2015robust,lovegrove2011accurate,mirabdollah2015fast,pereira2017monocular,scaramuzza2009absolute,yang2019cubeslam}, object size~\cite{botterill2011bag,botterill2012correcting,song2015high}, and pedestrian height~\cite{botterill2012correcting}, and estimate the absolute scale in monocular SLAM.
To reduce scale drift, \cite{strasdat2010scale} proposed a loop detection method that recognizes revisiting prior locations.
However, many driving trajectories do not complete a loop.
\cite{zhou2019ground} uses a camera height to estimate the absolute scale and correct scale drifts.
The method estimates the ground plane from a fixed region of interest (ROI) that corresponds to road area(s).
The estimation performance is limited if an ROI is blocked by obstacles such as preceding vehicles.
To resolve this limitation, this work uses road areas labeled by semantic segmentation.

To improve mapping accuracy, researchers combined monocular SLAM with deep learning-based semantic segmentation.
\cite{an2017semantic,brasch2018semantic} use semantic segmentation to probabilistically classify static and dynamic areas.
The Mask-SLAM method excludes dynamic objects and sky area, by using a mask produced by semantic segmentation \cite{kaneko2018mask}.
\cite{yu2018ds}~combines semantic segmentation with a moving consistency-check method to deal with dynamic objects.
These approaches showed excellent performance even for scenes with dynamic objects.
These methods, however, use a few segmented labels including dynamic objects and sky areas.
One may use more labels, such as road areas, to estimate ground planes.
More importantly, deep learning-based semantic segmentation has high computational costs, and applying it to each frame could terminate monocular SLAM. 
This limits the practical use of this powerful combination in real-time applications such as autonomous driving.
Different from existing methods~\cite{kaneko2018mask, an2017semantic, yu2018ds, brasch2018semantic}, 
this work applies semantic segmentation only to keyframes to achieve real-time monocular SLAM.

\subsection{Contributions and organization of the paper}

This paper proposes an improved real-time monocular SLAM framework that solves the aforementioned limitations of monocular SLAM, by using deep learning-based semantic segmentation in sophisticated ways.
The main contributions of this paper are as follows:
\begin{itemize}
\item
To resolve the time complexity problem of segmentation-based monocular SLAM, we apply deep learning-based semantic segmentation \textit{only} to keyframes.
To further improve the efficiency, segmentation process is performed in a different thread in parallel with mapping processes.
\item
To overcome the scale estimation challenges, we estimate the absolute scale using segmented road labels and the actual camera height, and apply it to correct scales of camera poses and 3D points.
\item
To reduce inaccurate mappings, we filter out, i.e., refine, corner features of both moving objects, such as cars, pedestrians, etc., and low-parallax areas, that may generate inappropriate 3D points. To discriminate low-parallax areas, we use the Kanade–Lucas–Tomasi (KLT) tracker~\cite{baker2004lucas} that is widely used in the optical flow, and select corner features with small displacement in the sky, terrain, and building areas.
\item
The proposed monocular SLAM framework using semantic segmentation simultaneously resolves the aforementioned scale estimation and mapping challenges.
\end{itemize}
Fig.~\ref{fig_intermediate_result} shows some intermediate results of the proposed monocular SLAM framework.

\begin{figure*}
\centering
\includegraphics[width=0.9\linewidth]{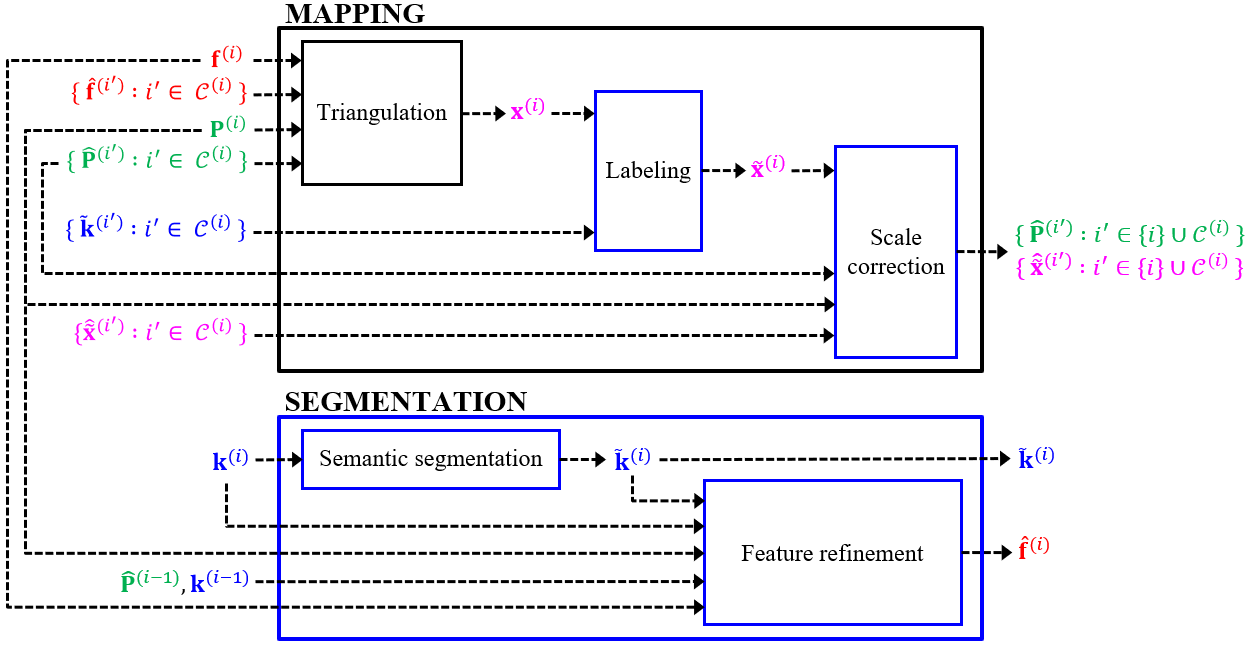}
\caption{Proposed combination of keyframe-based monocular SLAM and semantic segmentation. \BLUE{Blue boxes} indicate modified or added (sub-)modules.
\RED{Red symbols} indicate corner features, \GREEN{green symbols} indicate camera pose matrix, \BLUE{blue symbols} indicate keyframes, and \MG{magenta symbols} indicate 3D point map.
The input and output data sets are listed on left and right, respectively.}
\label{fig_method1}
\end{figure*}

The remainder of this paper is organized as follows.
Section~\ref{sec_proposed_method} 
describes the proposed efficient integration approach of deep learning-based semantic segmentation with monocular SLAM,
and the proposed scale correction and inaccurate mapping reduction methods by labeling 3D points and 2D corner features labeled based on segmented keyframes, respectively.
Section~\ref{sec_result_discussion} reports experiments 
where the proposed monocular SLAM system achieves significantly improved and comparable trajectory tracking accuracy, compared to existing state-of-the-art \emph{monocular} SLAMs and \emph{stereo} SLAM, respectively.
Furthermore, Section~\ref{sec_result_discussion} shows that the proposed system significantly reduces tracking time per frame compared to existing integration of segmentation into monocular SLAM \cite{kaneko2018mask}, achieving real-time processing on a standard CPU potentially with a standard GPU support.
Section~\ref{sec:conclusion} concludes the study and includes some future works.

\section{Proposed method}
\label{sec_proposed_method}

The proposed monocular SLAM system consists of three modules, localization, mapping, and segmentation, where each module runs in different threads.
We modify the backbone of the conventional keyframe-based monocular SLAM method \cite{mur2015orb}, efficiently integrating segmentation.
We summarize processes of each module between previous and current keyframe time point.
The localization module selects a keyframe when both mapping and segmentation modules complete processing on a previous keyframe;
similar to \cite{mur2015orb}, it provides corner features and camera pose of a current keyframe
(extracting corner features of each frame and estimating its camera pose with 3D points of connected keyframes).
The mapping module generates new 3D points by using triangulation between current corner features and connected corner features \cite{mur2015orb},
and assigns segmented labels from connected segmented keyframes to new 3D points.
It then estimates a ground plane using only road-labeled 3D points and correct a scale of camera poses and 3D points using estimated ground plane and real camera height.
While the mapping module provides scale-corrected camera poses and 3D points, the segmentation module performs deep learning-based segmentation to a downsampled keyframe and refines keyframe's corner features by removing moving objects and low-parallax areas using segmentation result.
Fig.~\ref{fig_system_overview} overviews the proposed monocular SLAM system.

Section~\ref{ss_preliminary} describes notations for data in the proposed method.
Section~\ref{ss_efficient_integration} describes how we integrate deep learning-based segmentation with keyframe-based monocular SLAM.
Using segmented keyframe results,
Sections~\ref{ss_scale_correction} and \ref{ss_incorrect_mapping_reduction} describe details of scale correction in 3D mapping and removal of factors in a current keyframe that potentially cause inappropriate mapping, respectively.

\subsection{Preliminaries}
\label{ss_preliminary}

This section summarizes notations for data in the proposed method: 
\begin{itemize}

\item The superscript $(\cdot)^{(i)}$ denotes the $i\text{th}$ keyframes index. $(i)$ indicates the current keyframe time point.

\item $\tilde{(\cdot)}$ denotes segmented image or points with semantic labels.
$\hat{(\cdot)}$ denotes that input is refined or scale-corrected.

\item $\mb{k}^{(i)}$ denotes the $i\text{th}$ keyframe.
$\tilde{\mb{k}}$ denotes segmented keyframe.
Fig.~\ref{fig_intermediate_result}(a) shows a segmented keyframe.

\item $\mb{f}^{(i)}$ denotes a 2D image with ORB corner features~\cite{rublee2011orb} extracted from $\mb{k}^{(i)}$ via the localization module.
Fig.~\ref{fig_intermediate_result}(b) shows extracted corner features.
$\hat{\mb{f}}$ denotes refined corner features 
by removing factors that potentially cause inappropriate mapping,
via the segmentation module.

\item $\mb{P}^{(i)}$ denotes $3 \times 4$ normalized camera pose matrix~\cite{hartley2003multiple} of $\mb{k}^{(i)}$. 
Normalized camera pose matrix is defined by $\mb{P} \deleq [\mb{R}, \mb{t}]$, where $\mb{R}$ and $\mb{t}$ denote a 3D rotation matrix and translation vector, respectively. These are calculated by the localization module.
$\hat{\mb{P}}$ denotes scale-corrected normalized camera pose matrix,
where its translation vector is calculated by scale-corrected coordinate of the camera center.
A camera center coordinate is calculated by $\mb{c} = -\mb{R}^{-1} \mb{t}$ \cite{hartley2003multiple}, 
where $(\cdot)^{-1}$ indicates the inverse of a matrix.
$\hat{\mb{c}}$ denotes scale-corrected camera center coordinate calculated from $\hat{\mb{P}}$.

\item $\mb{x}^{(i)}$ denotes a 3D map with 3D points generated from $\mb{f}^{(i)}$ via the mapping module.
$\hat{\mb{x}}$ denotes a scale-corrected 3D map, 
where a scale of 3D point coordinate are corrected.
$\hat{\tilde{\mb{x}}}$ denotes a scale-corrected 3D map with semantic labels.
Fig.~\ref{fig_intermediate_result}(c) illustrates 3D points with semantic labels.

\item 
$\mathcal{C}^{(i)}$ denotes a set of indices for connected keyframes with the $i\text{th}$ keyframe, in previous keyframe time points. 
If two keyframes generate 3D points at equal to or greater than $15$ same coordinates, they are called \textit{connected} \cite{mur2015orb}.
Specifically, $\mathcal{C}^{(i)} := \{ i-1, \ldots, i-C^{(i)} \}$, where $C^{(i)}$ is the number of connected keyframes with $\mb{k}^{(i)}$.

\end{itemize}

\subsection{Proposed efficient integration of semantic segmentation with monocular SLAM}
\label{ss_efficient_integration}

To efficiently integrate semantic segmentation with monocular SLAM, we apply semantic segmentation only to keyframes.
Applying semantic segmentation to monocular SLAM will improve the monocular SLAM accuracy by preventing generating inappropriate 3D points \cite{kaneko2018mask,an2017semantic,yu2018ds,brasch2018semantic} and assigning semantic labels to 3D points for better ground-plane estimation.
However, applying semantic segmentation to every frame is computational demanding and unsuitable for real-time monocular SLAM.
3D points are generated from the mapping process that uses only keyframes, so segmenting only keyframes is sufficient for processing 3D points in monocular SLAM.

\begin{figure}
\centering
\includegraphics[width=0.95\linewidth]{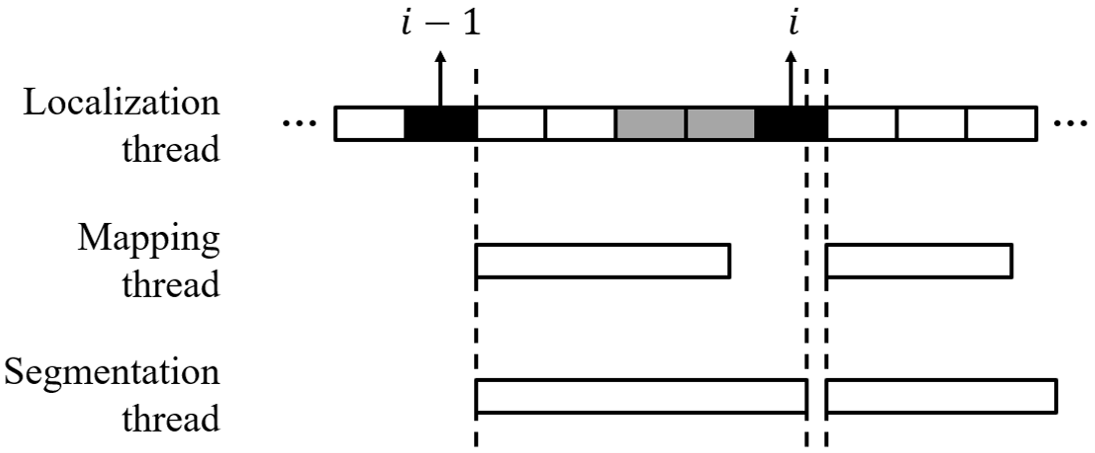}
\caption{
Proposed keyframe selection strategy and its timing. 
In localization thread, the length of each box indicates the processing time for each frame.
In mapping and segmentation thread, 
the length of each box indicates the processing time for each keyframe.
Gray and black box denotes keyframe candidate and selected keyframe, respectively.
We choose a keyframe only after both mapping and segmentation processes are completed,
although localization thread flags frame(s) as keyframe candidate(s).
Each candidate is determined at the end of each frame localization.
}
\label{fig_method1_2}
\end{figure}

In addition,
we perform the mapping and segmentation process in two different threads.
In monocular SLAM, 3D points are generated by triangulating matched features from two keyframes~\cite{hartley2003multiple}.
Two keyframes have matched features, so segmenting only one of two keyframes may be sufficient for subsequent 3D point processing.
The proposed integration approach 
performs the mapping process with connected segmented keyframes and related data in one thread,
in parallel with performing the segmentation process with a current keyframe and related data in the other thread.
Fig.~\ref{fig_method1} illustrates the parallel processing.
We describe details of each thread processing below.

In the mapping thread, we first generate a 3D point map $\mb{x}^{(i)}$ by triangulation between current corner features $\mb{f}^{(i)}$ and each of connected (refined) corner features $\{ \hat{\mb{f}}^{(i')} : i' \in \mathcal{C}^{(i)} \}$, using a current camera matrix $\mb{P}^{(i)}$ and each of connected (scale-corrected) camera matrices $\{ \hat{\mb{P}}^{(i')} : i' \in \mathcal{C}^{(i)} \}$.
We then assign semantic labels in $\{ \tilde{\mb{k}}^{(i')} : i' \in \mathcal{C}^{(i)} \}$ to generated $\mb{x}^{(i)}$ and obtain $\tilde{\mb{x}}^{(i)}$.
Finally, we correct a scale of keyframe's camera center coordinates computed by each of $\{ \mb{P}^{(i)}, \hat{\mb{P}}^{(i')} : i' \in \mathcal{C}^{(i)} \}$ and that of 3D point coordinates in each of $\{ \tilde{\mb{x}}^{(i)}, \hat{\tilde{\mb{x}}}^{(i')} : i' \in \mathcal{C}^{(i)} \}$.
While the mapping thread performs aforementioned processes,
the segmentation thread first segments a current keyframe $\mb{k}^{(i)}$, and obtains $\tilde{\mb{k}}^{(i)}$. 
We then refine a current corner feature $\mb{f}^{(i)}$, using current data $\{ \mb{P}^{(i)}, \mb{k}^{(i)}, \tilde{\mb{k}}^{(i)} \}$ and previous data $\{ \hat{\mb{P}}^{(i-1)}, \mb{k}^{(i-1)} \}$, and obtain $\hat{\mb{f}}^{(i)}$.
Fig.~\ref{fig_method1} summarizes the above processes.
We will describe details of the scale correction and corner feature refinement process later in Section~\ref{ss_scale_correction} 
and \ref{ss_incorrect_mapping_reduction}, respectively.

Note that we choose a keyframe only after both the mapping and segmentation processes are completed.
If the localization thread generates keyframe candidates in the middle of mapping or segmentation process, we disregard them and choose the one generated right after the mapping and segmentation processes are completed as a keyframe.
(It is likely that subsequent frames after a keyframe candidate are also keyframe candidates.)
Fig.~\ref{fig_method1_2} summarizes this keyframe selection strategy.
Specifically, 
we select $\{ \mb{f}^{(i)}, \mb{P}^{(i)}, \mb{k}^{(i)} \}$ from the localization module and begin the mapping and segmentation processes, 
only after $\{ \hat{\mb{P}}^{(i-1)}, \hat{\tilde{\mb{x}}}^{(i-1)} \}$ and
$\{ \tilde{\mb{k}}^{(i-1)}, \hat{\mb{f}}^{(i-1)} \}$ are
obtained by the mapping and segmentation thread in the previous keyframe time point, respectively.

Lastly, we downsample each 2D keyframe to reduce computational complexity of semantic segmentation (and upsample segmented keyframes back to the original resolution).
Considering the above keyframe selection strategy and observations that semantic segmentation is generally more computationally expensive than the mapping module,
the segmentation computation time critically affects the decision timing of a new keyframe.
The slower semantic segmentation, the longer time gap between keyframes. 
This leads to fewer matched corner features between two keyframe time points and fewer generated 3D points via triangulation.
The localization module projects 3D points onto each frame and estimates camera pose from projected 3D points and extracted corner features \cite{mur2015orb}.
If the number of 3D points is small, then pose estimation accuracy reduces and eventually monocular SLAM performance significantly degrades.
Reducing the keyframe image size/resolution can decrease the computational costs of semantic segmentation methods using convolutional neural network (CNN), at a cost of segmentation accuracy.
Our experiments disregarding segmentation CNN computation costs demonstrate that although CNN segmentation accuracy decreases by reduced image size, the proposed monocular SLAM gives comparable performance with that using the original image size. 
See details later in Section~\ref{exp_decision_resize_factor}.

\subsection{Scale correction}
\label{ss_scale_correction}

At each current keyframe time point, 
we correct the scale of keyframe's camera center coordinate and each 3D point coordinate.
We first estimate ground plane parameters $\{ \mb{n}^{(i)}, d^{(i)} \}$ using only \emph{road} labeled 3D points in $\{ \tilde{\mb{x}}^{(i)}, \hat{\tilde{\mb{x}}}^{(i')} : i' \in \mathcal{C}^{(i)} \}$.
A ground plane is parameterized with a normal vector $\mb{n}^{(i)}$ ($\| \mb{n} \|_2 = 1$) and the perpendicular distance between a ground plane and the origin, $d^{(i)}$, using the plane equation in 3D $\langle \mb{n}, (x, y, z) \rangle 
= 0$ \cite{anton2013elementary},
where $\langle \cdot, \cdot \rangle$ denotes the inner product between two vectors
and $(x,y,z)$ is a coordinate of any 3D point on a plane.
For accurate ground plane estimation,
we use a robust parameter estimation method,
random sample consensus (RANSAC) \cite{fischler1981random} that uses only inliers of road-labeled 3D points
(we normalized $\mb{n}$ at each RANSAC iteration).
Different from existing methods that use unlabeled 3D points in ROI for RANSAC, e.g., \cite{zhou2019ground},
we use 3D points with road label.

We then compute a scaling factor~\cite{grater2015robust} using
\emph{virtual} camera height and a \emph{real} camera height $h_{\text{real}}$ that is measured in advance.
We estimate a \emph{virtual} camera height $h^{(i)}$ with the estimated ground plane parameters $\{ \mb{n}^{(i)}, d^{(i)} \}$ above.
A \emph{virtual} camera height $h^{(i)}$ is the perpendicular distance between estimated ground plane and a current keyframe's camera center coordinate $\mb{c}^{(i)}$, and calculated by $h^{(i)} = | \langle \mb{n}^{(i)}, \mb{c}^{(i)} \rangle + d^{(i)} |$.
We compute a scaling factor $s^{(i)}$ using $h_{\text{real}}$ and $h^{(i)}$: 
\begin{equation}
\label{eq:s^i}
s^{(i)} = \frac{h_{\text{real}}}{h^{(i)}}.
\end{equation}
Intuitively speaking, $s^{(i)}$ above quantifies the scale drift at the current keyframe time point.

Finally, we adjust coordinates of keyframe's camera center computed by each of $\{ \mb{P}^{(i)}, \hat{\mb{P}}^{(i')} : i' \in \mathcal{C}^{(i)} \}$ and that of each 3D point in $\{ \tilde{\mb{x}}^{(i)}, \hat{\tilde{\mb{x}}}^{(i')} : i' \in \mathcal{C}^{(i)} \}$, using the scaling factor $s^{(i)}$ in \R{eq:s^i}.
Applying $s^{(i)}$ follows the scale correction processes in \cite{zhou2019ground}.
We perform scale correction when 
the scaling factor satisfies $0.001 < \left | s^{(i)} - 1 \right | < 0.2$, to avoid frequent or intensive correction.
We adjust camera center coordinates of keyframes, 
by scaling the distance between $\mb{c}^{(i)}$ and a camera center coordinate calculated from each of $\{ \hat{\mb{P}}^{(i')} : i' \in \mathcal{C}^{(i)} \}$ with $s^{(i)}$.
Likewise, we adjust 3D point coordinates by scaling the distance between $\mb{c}^{(i)}$ and each 3D point coordinate in $\{ \hat{\tilde{\mb{x}}}^{(i')} : i' \in \mathcal{C}^{(i)} \}$ with $s^{(i)}$.

In the first several keyframes, we do not perform the scale correction 
since road-labeled 3D points are insufficient.
When the number of road-labeled 3D points become large or equal than $50$,
we estimate $h^{(i)}$ by averaging the absolute differences between $\mb{c}^{(i)}$'s $y$-coordinate and $y$-coordinates of road-labeled 3D points in $\{ \tilde{\mb{x}}^{(i')} : i' \in \{ i \} \cup \mathcal{C}^{(i)} \}$, and correct scales of $\{ \tilde{\mb{x}}^{(i')}, \mb{P}^{(i')} : i' \in \{ i \} \cup \mathcal{C}^{(i)} \}$ with $s^{(i)}$ in \R{eq:s^i}.
(RANSAC uses a user-specific threshold in determining inlier samples and this threshold depends on scales of samples \cite{grater2015robust}, 
but such 3D points are not yet scale-corrected in this keyframe time point.)
Subsequent keyframes will use RANSAC-based ground plane estimation for the scale correction process, as described earlier.

\begin{figure}[t!]
\centering
\begin{tabular}{P{8.5cm}}
\includegraphics[width=1\linewidth]{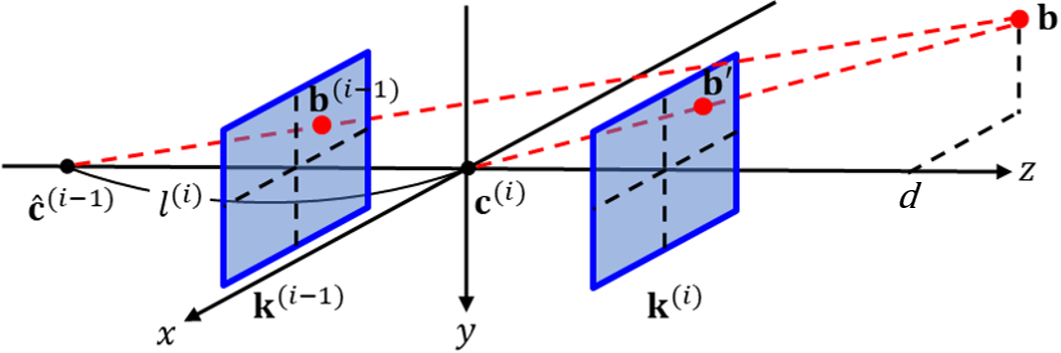}
\\
{\footnotesize (a) 3D pin-hole camera projection on current and previous keyframes.\vspace{0.5em}}
\\
\includegraphics[scale=0.41]{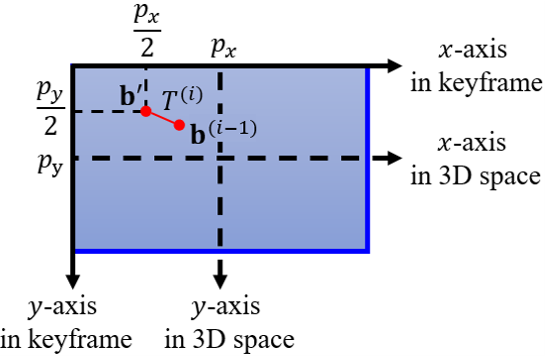} 
\\
{\footnotesize (b) $T^{(i)}$ is the distance between $\mb{b}'$ and $\mb{b}^{(i-1)}$ on 2D keyframe plane.}
\\
\vspace{-1.15em}
{\footnotesize The principal point coordinate $(p_x, p_y)$ is defined on 2D keyframe plane.}
\end{tabular}
\caption{
Geometrical illustrations of 
the proposed adaptive $T^{(i)}$ selection using 3D pin-hole projection model on current and previous keyframes.
}
\label{fig_method3}
\end{figure}

\subsection{
Corner feature refinement by removing inappropriate mapping factors
}
\label{ss_incorrect_mapping_reduction}

At each current keyframe time point,
we refine current corner features $\mb{f}^{(i)}$ by removing corner features of moving objects and low-parallax areas, using current data $\{ \mb{P}^{(i)}, \mb{k}^{(i)}, \tilde{\mb{k}}^{(i)} \}$ and previous data $\{ \hat{\mb{P}}^{(i-1)}, \mb{k}^{(i-1)} \}$.
Using a current segmented keyframe $\tilde{\mb{k}}^{(i)}$,
we first remove corner features in areas labeled as movable objects, such as car, person, and bicycle, etc., similar to the existing methods~\cite{kaneko2018mask,an2017semantic,brasch2018semantic}.
We next remove corner features labeled as sky, terrain, and building -- referred to as backgrounds -- in low-parallax areas.
For each corner feature in backgrounds, we compute the displacement between a current keyframe $\mb{k}^{(i)}$ and a previous keyframe $\mb{k}^{(i-1)}$, using KLT tracker~\cite{baker2004lucas}.
Corner features with the displacement less than some threshold are considered as low-parallax areas.

When a 3D point is projected on a current keyframe and a previous keyframe,
the displacement between projected 2D points changes depending on the distance between two keyframes.
We adaptively select a threshold $T^{(i)}$ at each current keyframe time point, using the 3D projection model on two keyframes (see below) and estimated distance between current and previous keyframes.

This paragraph describes the 3D pin-hole camera projection model on current and previous keyframes, and its notations.
A 3D point with coordinate $\mb{b}$ projects 
onto $\mb{k}^{(i)}$- and $\mb{k}^{(i-1)}$-keyframe 2D planes 
(towards their camera centers with coordinates $\mb{c}^{(i)}$ and $\hat{\mb{c}}^{(i-1)}$, respectively).
The mapped points on current and previous keyframes are denoted by $\mb{b}'$ and $\mb{b}^{(i-1)}$, respectively.
The distance between $\mb{c}^{(i)}$ and $\hat{\mb{c}}^{(i-1)}$ along the $z$-direction is denoted by $l^{(i)}$. 
Fig.~\ref{fig_method3}(a) illustrates this.
In setting or estimating $\mb{b}$, $\mb{b}'$, and $\mb{b}^{(i-1)}$,
we use camera intrinsic parameters, focal length $f$ and principal point coordinate $(p_x, p_y)$ (obtained by camera calibration in advance). 

The proposed adaptive $T^{(i)}$ selection consists of four steps.
First, we set $\mb{b}'$ using principal point coordinate:
\begin{equation}
\label{eq:b'}
\mb{b}' = 
\begin{bmatrix}
\frac{p_{x}}{2}
\\ 
\frac{p_{y}}{2}
\end{bmatrix},
\end{equation}
where we observed that in general, low-parallax areas are located near this coordinate; see its location on a 2D keyframe plane in Fig.~\ref{fig_method3}(b).
Second, we set $\mb{b}$ as follows:
we set the $z$-coordinate of $\mb{b}$ as $d$ by assuming that inappropriate mapping occurs in objects at a distance of $d$ m or more and calculate $x$- and $y$-coordinates of $\mb{b}$ by
\begin{equation}
\label{eq:b}
\mb{b} = 
\begin{bmatrix}
- \frac{d p_{x}}{2f}
\\ 
- \frac{d p_{y}}{2f}
\\
d
\end{bmatrix}.
\end{equation}
From this perspective, we refer to $d$ as the low-parallax feature determination parameter.
We obtained the result in \R{eq:b} by back-projecting $\mb{b}'$ in \R{eq:b'} to $\mb{b}$ using the perspective projection equation~\cite{hartley2003multiple}:
\begin{equation*}
\lambda
\begin{bmatrix}
\mb{b}'\\ 
1
\end{bmatrix}
=
\begin{bmatrix}
f &0  &p_{x} \\ 
0 &f  &p_{y} \\ 
0 &0  &1 
\end{bmatrix}
\mb{b},
\end{equation*}
where we set $\lambda = d$ to express $\mb{b}'$ in the form of homogeneous coordinate.
Here, the $3\times3$ matrix is the camera intrinsic matrix.
Third, we estimate $\mb{b}^{(i-1)}$ using $l^{(i)}$:
\begin{equation}
\label{eq:b^i-1}
\mb{b}^{(i-1)} = 
\begin{bmatrix}
\frac{p_{x} (l^{(i)}+d/2)}{l^{(i)}+d}\\
\frac{p_{y} (l^{(i)}+d/2)}{l^{(i)}+d}
\end{bmatrix}.
\end{equation}
We obtained the result in \R{eq:b^i-1} by
mapping $\mb{b}$ in \R{eq:b} with the $z$-coordinate increased by $l^{(i)}$ onto $\mb{k}^{(i-1)}$-keyframe 2D plane
(using the perspective projection):
\begin{equation*}
\lambda
\begin{bmatrix}
\mb{b}^{(i-1)}\\ 
1
\end{bmatrix}
=
\begin{bmatrix}
f &0  &p_{x} \\ 
0 &f  &p_{y} \\ 
0 &0  &1 
\end{bmatrix}
\begin{bmatrix}
\frac{- d p_{x}}{2f}\\ 
\frac{- d p_{y}}{2f}\\
l^{(i)}+d
\end{bmatrix}
,
\end{equation*}
where we set $\lambda = l^{(i)}+d$ to express $\mb{b}^{(i-1)}$ in the form of homogeneous coordinate.
Finally, we select a threshold $T^{(i)}$ by calculating the distance between $\mb{b}'$ in \R{eq:b'} and $\mb{b}^{(i-1)}$ in \R{eq:b^i-1}:
\begin{equation}
\label{eq:T^(i)}
T^{(i)} = \| \mb{b}'-\mb{b}^{(i-1)} \|_2 = \frac{l^{(i)}}{2(l^{(i)}+d)}\sqrt{p_{x}^{2}+p_{y}^{2}}.
\end{equation}
This is illustrated in Fig.~\ref{fig_method3}(b).
Objects more than $d$ m away from the $\mb{c}^{(i)}$-point may have the displacement less than $T^{(i)}$. We remove the corresponding corner features with displacement less than $T^{(i)}$.
We determine an appropriate low-parallax feature determination parameter $d$ by comparing performances of proposed monocular SLAM with different $d$ values (see details later in \ref{experiment_different_low_parallax_feature_factor}).

\section{Results and discussion}
\label{sec_result_discussion}

We compared the proposed monocular SLAM with 
four variations of ORB-SLAM2 \cite{mur2017orb}: 
\textit{1)} monocular setup without loop closing (LC), 
\textit{2)} monocular setup with LC, 
\textit{3)} stereo setup without LC,
and
\textit{4)} stereo setup with LC.
We additionally compared the proposed method with Mask-SLAM \cite{kaneko2018mask}, a representative monocular SLAM method that uses semantic segmentation.
The ORB-SLAM2 stereo setup uses two cameras and 
resolves the scale ambiguity \& drift~\cite{mur2017orb} and
inaccurate mapping issues \cite{scaramuzza2011visual}.
LC can further improve the estimation accuracy of camera poses and 3D point coordinates in ORB-SLAM2,
if it detects a loop.
It can be applied to both monocular and stereo setups.
Mask-SLAM applies semantic segmentation to all frames to exclude dynamic objects.
We mainly focus on comparing the trajectory tracking accuracy between different visual SLAMs that do \emph{not} experience tracking failure, 
where tracking failure may additionally need relocalization \cite{eade2008unified,williams2011automatic,mur2014fast}.

In measuring the trajectory tracking accuracy,
we calculated absolute trajectory error (ATE) \cite{sturm2012benchmark}, root-mean-squared-error between
predicted camera center coordinates of keyframes and their ground-truth in meters (m) (if ground-truth exists).
Throughout experiments, we did not use the manual re-alignment process \cite[Sec. VIII-E]{mur2015orb} that realigns an estimated trajectory to the ground-truth, since this is impractical.

\subsection{Experimental setup}
\label{ss_experimental_setup}

We first determined an appropriate downsampling factor for keyframe segmentation (see the last paragraph in Section~\ref{ss_efficient_integration}),
and compared the proposed method using the determined downsampling factor with the five state-of-the-art visual SLAM methods described above.
We used either a GPU or CPU for semantic segmentation of keyframe images, and ran remaining modules on a CPU.
We used 3.1 GHz Intel Core i9 with 128 GB RAM and 1607 MHz NVIDIA GeForce GTX 1080 with 8GB RAM, for CPU processing and GPU processing, respectively.

In the localization module,
we extracted $3,\!000$ ORB corner features \cite{rublee2011orb} for each input frame.

\subsubsection{Comparisons of segmentation and tracking accuracies and segmentation running time, with different keyframe (spatial) resolutions}
\label{experiment_comparison_resolutions}

We compared the accuracy of semantic segmentation and the trajectory tracking and running time of semantic segmentation of the proposed monocular SLAM system with different keyframe resolutions.
To generate keyframes with different resolutions, we downsampled the width and height of original keyframes by factors of $4/3$, $2$, or $4$ using bicubic dowmsampling.
We upsampled segmented keyframes with nearest neighbor interpolation (using an upsampling factor identical to corresponding downsampling factor).
In the proposed method, long segmentation running time(s) can jeopardize camera pose estimation in localization (due to limited 3D points), so we disregarded segmentation computation costs in measuring trajectory tracking accuracy.

For keyframe segmentation with a GPU support, we used the advanced NVIDIA semantic segmentation network \cite{zhu2019improving}.\footnote{This method uses video prediction based data synthesis \cite{reda2018sdc} to increase the training data size, and it took the first place in the KITTI semantic segmentation benchmark \cite{andreas2020kitti_semantic}.}
This network was pre-trained with the KITTI semantic segmentation dataset \cite{Alhaija2018IJCV} that is different from the KITTI odometry dataset \cite{Geiger2012CVPR}.
The proposed monocular SLAM method uses four classes, road, movable objects, background, and others,
where we categorized predictions of the segmentation network as follows:  
road (``road''), 
movable objects (``person'', ``rider'', ``car'', ``truck'', ``bus'', ``train'', ``motorcycle'', and ``bicycle''),
background (``building'', ``terrain'', and ``sky''), 
and others.

For CPU-based keyframe segmentation, we used ERFNet \cite{romera2017erfnet} that can achieve high computational efficiency by using residual connections and factorized convolutions.
Because ERFNet is designed to use input with the specific aspect ratio ($\text{width} : \text{height} = 2:1$), we adjusted the size of keyframe images to fit the ratio. (We only tested the factor 1 and $4/3$ for ERFNet.)
This network was also trained with the KITTI semantic segmentation dataset \cite{andreas2020kitti_semantic}, with hyperparameters given in \cite{romera2017erfnet} except for $200$ epochs and a batch size of $6$.
We categorized predictions of ERFNet into the aforementioned four classes used in the NVIDIA network.

In measuring the trajectory tracking accuracy and segmentation running time, 
we used the sequence $00$ of the KITTI odometry dataset~\cite{Geiger2012CVPR}.
In measuring the segmentation accuracy, we used the Cityscapes dataset~\cite{Cordts2016Cityscapes} that consists of $3,475$ images driving in urban environments, since the KITTI-$00$ dataset does not have ground-truth segmentation annotations.
We evaluated the semantic segmentation accuracy by mean intersection over union (IoU) of the road, movable objects, and background labels (others were not used in the proposed method).

We determined an appropriate spatial downsampling factor for keyframe segmentation considering trade-off between trajectory tracking accuracy and segmentation running time.

\begin{table*}[t!]
\caption{Segmentation accuracy, running time and trajectory tracking accuracy with different processing methods and different keyframe downsampling factors in proposed monocular SLAM (KITTI-00 dataset)}
\begin{center}
\begin{tabular}{c|c|ccc|rr|r}
\hline\hline
\multirow{2}{*}{\begin{tabular}[c]{@{}c@{}}Segmentation\\ methods\end{tabular}} & \multirow{2}{*}{\begin{tabular}[c]{@{}c@{}}Downsampling\\ factor\\ (spatial resolution)\end{tabular}} & \multicolumn{3}{c|}{\begin{tabular}[c]{@{}c@{}}Segmentation\\ accuracy (IoU)\end{tabular}} & \multicolumn{2}{c|}{\begin{tabular}[c]{@{}c@{}}Segmentation\\ running\\ time (ms)\end{tabular}} & \multirow{2}{*}{\begin{tabular}[c]{@{}c@{}}ATE\\ (m)\end{tabular}} \\ \cline{3-7} & & 
Road  & \begin{tabular}[c]{@{}c@{}}Movable\\ obj.\end{tabular} & Background & Mean & Std. & \\ \hline
\multirow{4}{*}{NVIDIA network on GPU} & 
1 (1241$\times$376) & 0.934 & 0.865 & 0.867 & 479.05 & 6.72 & 6.721 \\ & 
$4/3$ (930$\times$282) & 0.926 & 0.863 & 0.863 & 288.44 & 5.90 & 7.755 \\ &
2 (620$\times$188) & 0.918 & 0.859 & 0.855 & 129.44 & 4.13 & 6.784 \\ & 
4 (310$\times$94) & 0.898 & 0.782 & 0.806 & 61.66 & 4.94 & 8.681 \\ \hline
\multirow{2}{*}{ERFNet on CPU} & 
$\approx$1 (960$\times$480) & 0.912 & 0.627 & 0.825 & 238.90 & 13.66 & 8.031 \\ &
$4/3$ (720$\times$360) & 0.877 & 0.551 & 0.782 & 131.99 & 22.20 & 11.932 \\
\hline\hline
\end{tabular}
\end{center}
\label{segmentation_comparison}
\end{table*}

\subsubsection{Comparisons of proposed monocular SLAM with different sub-module combinations}
\label{sss_module_combinations}

We ran the proposed monocular SLAM system with different low-parallax feature determination parameters $d = 200, 250, 300$ and chose an appropriate one by considering both the trajectory tracking accuracy and tracking success rate.
We used the chosen parameter throughout the experiments.

In addition, we ran an ablation study by removing the scale correction and feature refining sub-modules, respectively.
We measured the trajectory tracking accuracy from each setup.

Throughout the experiments in this section,
we used the GPU-based NVIDIA network with a determined keyframe resolution in Section~\ref{experiment_comparison_resolutions}, and the sequence $00$ of the KITTI odometry dataset~\cite{Geiger2012CVPR}.

\subsubsection{Comparisons between different visual SLAM methods} 
\label{experiment_comparison_vslam}

We ran the proposed monocular SLAM considering segmentation computation times with a determined keyframe spatial resolution via the trade-off investigation experiment in Section~\ref{exp_decision_resize_factor}.
We tested both segmentation processing methods, the NVIDIA network on a GPU and ERFNet on a CPU.
Segmentation running time includes computational costs of downsampling, semantic segmentation, upsampling, and data transfer times between CPU and GPU (if GPU is used).
We segmented all downsampled frames in advance; when a keyframe is selected,
we added the corresponding segmentation running time to the segmentation thread.

For comparing the trajectory tracking accuracy with different visual SLAM methods, we used the sequences $00$, $02$, $03$, $04$, $05$, $07$, and $08$ of the KITTI odometry dataset \cite{Geiger2012CVPR}.
The KITTI odometry dataset consists of sequence sets, where each sequence set is obtained by driving in a different residential area.
The original image size is $1241\times376$ and
the camera frame-rate is $10$ Hz.
The sequences $00$, $02$, $05$, and $07$ have a loop or loops in their trajectories.
The sequences $03$, $04$, and $08$ do not have loop(s),
so we only ran ORB-SLAM2 without LC (for both monocular and stereo setups).
To consider dynamic environments with moving objects, we used the KITTI 02 and 04 sequences.

In addition, we used a highway sequence provided by StradVision Inc.~\cite{stradvision_2021} to consider environments with both moving objects and distant objects. 
The StradVision sequence includes monocular video of $900$ frames where many cars drive on the highway (i.e., there exist many moving objects) surrounded by mountains (i.e., low-parallax area).
The StradVision dataset was obtained with imaging resolution $1280\times720$ and camera frame-rate $30$ Hz.
In experiments with the StradVision dataset, we compared the proposed method using the NVIDIA network, with monocular ORB-SLAM2.
The sequence does not have any loops, so we only ran monocular ORB-SLAM2 without LC.

In addition to the quantitative analyses with ATEs, we qualitatively compared trajectory tracking accuracy in the KITTI odometry dataset by comparing estimated trajectories with their ground-truths.
Since the StradVision sequence does not have ground-truth trajectory, we aligned trajectories with the map provided by Google Maps for qualitative comparison.
(In monocular ORB-SLAM2 without LC, we also aligned a scale of the estimated trajectory.)

To study the feasibility of running semantic segmentation with monocular SLAM in real-time, we calculated the tracking time per frame in milliseconds (ms).
We compared the proposed efficient combination framework of monocular SLAM and semantic segmentation, with the existing integration, Mask-SLAM.
In measuring the tracking time per frame, we used all the aforementioned KITTI sequences and calculated its average and standard deviation values.

\subsection{Comparisons of segmentation and tracking accuracies and segmentation running time, with different keyframe (spatial) resolutions}
\label{exp_decision_resize_factor}

\subsubsection{Segmentation accuracy comparisons with different keyframe resolutions}

For GPU-based semantic segmentation with the NVIDIA network,
Table \ref{segmentation_comparison} shows that when keyframes are downsampled by a factor of $4/3$ or $2$, segmentation accuracy (slightly) decreases, compared to the case with the original keyframe resolution.
When keyframes are downsampled by a factor of $4$, the segmentation accuracy significantly decreases, compared to the case with the original keyframe resolution.
Specifically, the segmentation accuracy (in IoU) decreases by $3.85\%$, $9.60\%$, and $7.04\%$ in road, movable objects, and background labels, respectively.
Comparing the results of the NVIDIA network and ERFNet without downsampling in Table \ref{segmentation_comparison} shows that ERFNet running on CPU slightly degrades the segmentation accuracy of the NVIDIA network on GPU. 
The trend applies to comparisons with NVIDIA segmentation network (on GPU) using a downsampling factor up to 2; see the NVIDIA network results in Table.
Comparing the downsampling factor $4/3$ results between the NVIDIA network and ERFNet in Table \ref{segmentation_comparison} shows that ERFNet on CPU significantly degrades the segmentation accuracy, compared to the NVIDIA network on GPU.

\subsubsection{Tracking accuracy comparisons with different keyframe resolutions}
\label{experiment_comparison_segmentation_accuracy_resolutions}

For GPU-based semantic segmentation with NVIDIA network, Table \ref{segmentation_comparison} shows that although keyframes are downsampled by a factor of $4/3$ or $2$, the trajectory tracking accuracy is similar to the case with the original keyframe resolution.
When keyframes are downsampled by a factor of $4$, trajectory tracking accuracy significantly decreases, compared to the case without downsampling.
Comparing the results of the NVIDIA network and ERFNet without downsampling in Table~\ref{segmentation_comparison} shows that using ERFNet running on CPU decreases the tracking accuracy of the proposed method using the NVIDIA network on GPU.
The trend applies to comparisons with NVIDIA network segmentation (on GPU) using a downsampling factor up to 2; see the NVIDIA network results in Table~\ref{segmentation_comparison}.
Comparing the downsampling factor $4/3$ results between the NVIDIA network and ERFNet in Table \ref{segmentation_comparison} shows that ERFNet on CPU significantly degrades the trajectory tracking accuracy.

\subsubsection{Segmentation running time comparisons with different keyframe resolutions}

Table \ref{segmentation_comparison} shows that segmentation running time approximately linearly decreases, as the keyframe downsampling factor increases, regardless of network and processing unit choices. 
This is natural, as the computational complexity of segmentation network scales with the size of input images.
Note that CPU-based ERFNet without downsampling runs segmentation faster then the GPU-based NVIDIA network using the downsampling factor $4/3$. Compare the corresponding results in Table \ref{segmentation_comparison}.

\subsubsection{Real-time processing feasibility of the proposed system with different keyframe resolutions in segmentation}

When segmentation computation times are considered (see Section~\ref{experiment_comparison_vslam}), 
we observed with the KITTI-$00$ sequence that
the proposed monocular SLAM system runs continuously with the GPU-based NVIDIA networks using the downsampling factors 2 and 4, and CPU-based ERFNets using no downsampling and the downsampling factor $4/3$; with the downsampling factors 1 and $4/3$ in GPU-based NVIDIA net, the proposed system frequently stops running.
The reason is that long segmentation running time can significantly delay keyframe selection and limit 3D point generation, and eventually, cause the entire system failure;
see further details in Section~\ref{ss_efficient_integration}.
We observed that if the NVIDIA network runs on a standard CPU, the proposed system eventually stops running even with the keyframe downsampling factor 4 for segmentation.

To achieve real-time execution of the proposed monocular visual SLAM system
on the test datasets,
we decide to spatially downsample keyframes by the factor $2$ 
in the GPU-based NVIDIA network and to use no downsampling in CPU-based ERFNet for semantic segmentation.
For the proposed system with NVIDIA segmentation network running on a standard GPU, the downsampling factor 2 for segmentation shows a reasonable trade-off between tracking accuracy and segmentation running time. 
This is supported by the empirical observations in Section~\ref{exp_decision_resize_factor}.
The keyframe downsampling factor 2 for segmentation \textit{1)} slightly degrades the trajectory tracking accuracy (only by $0.94\%$), but \textit{2)} significantly reduces the segmentation running time (by a factor of $\approx 4$), so proposed monocular SLAM runs continuously.
For the proposed system with ERFNet running a standard CPU, we decide not to use downsampling for keyframes in segmentation, because even with the downsampling factor $4/3$, the trajectory tracking accuracy significantly degrades.

One may have a different favorable keyframe downsampling factor for other datasets.
We suggest to choose a keyframe downsampling factor (for segmentation) that achieves segmentation accuracy higher than $0.9$ IoU for road labels. 

\subsection{Comparisons of proposed monocular SLAM with different module combinations}
\label{ss_module_combinations}

\subsubsection{Comparisons of proposed monocular SLAM with different low-parallax feature determination parameters}
\label{experiment_different_low_parallax_feature_factor}

Table~\ref{different_low_parallax_feature_factor} shows that as the low-parallax feature determination parameter $d$ increases, the number of filtered corner features and the trajectory tracking accuracy decrease.
When $d$ is set to $200$, the trajectory tracking accuracy is slightly better compared to the $d=250$ case.
However, the proposed system frequently stops running, as shown by the low tracking success rate in Table~\ref{different_low_parallax_feature_factor}, where the tracking success rate denotes the ratio of number of frames in which localization is successfully executed, to total number of frames.
This is because too many corner features are removed, when $d=200$.
Therefore we set the low-parallax feature determination parameter to $250$ for the proposed corner feature refinement.

\begin{table}[t!]
\centering
\caption{Averaged number of filtered corner features, trajectory tracking accuracy, and tracking success rate with different low-parallax determination feature determination parameters in proposed monocular SLAM (KITTI-00 dataset)}
\begin{tabular}{c|c|c|c}
\hline\hline
\begin{tabular}[c]{@{}c@{}} Low-parallax feature \\ determination parameter \end{tabular} & \begin{tabular}[c]{@{}c@{}} Averaged number of \\ filtered corner features \end{tabular} & \begin{tabular}[c]{@{}c@{}} ATE \\ (m) \end{tabular} & \begin{tabular}[c]{@{}c@{}} Tracking \\ success \\ rate (\%) \end{tabular}\\ \hline
$200$ & 291.24 & 6.409 & 42.06 \\
$250$ & ~96.75 & 6.797 & 100 \\
$300$ & ~36.26 & 8.459 & 100 \\ 
\hline\hline
\end{tabular}
\label{different_low_parallax_feature_factor}
\end{table}

\begin{table}[t!]
\centering
\caption{Ablation study of scale correction and feature refinement sub-modules in proposed monocular SLAM (KITTI-00 dataset)}
\renewcommand{\tabcolsep}{0.9mm}
\begin{tabular}{c|cccc}
\hline\hline
 & \begin{tabular}[c]{@{}c@{}} (a) Proposed \\ mono. SLAM \\ w/ both \\ sub-modules \end{tabular} & \begin{tabular}[c]{@{}c@{}} (b) Proposed \\ mono. SLAM \\ w/ only scale \\ correction \end{tabular} & \begin{tabular}[c]{@{}c@{}} (c) Proposed \\ mono. SLAM \\ w/ only feature \\ refinement \end{tabular} & \begin{tabular}[c]{@{}c@{}} (d) Mono. \\ ORB-SLAM2 \\ w/o LC \end{tabular} \\ \hline
ATE (m) & \multicolumn{1}{r}{\textbf{6.797}} & \multicolumn{1}{r}{9.253} & \multicolumn{1}{r}{212.616} & \multicolumn{1}{r}{231.701} \\   
\hline\hline
\end{tabular}
\label{table_ablation_study}
\end{table}

\subsubsection{Ablation study of proposed monocular SLAM}
\label{sss_ablation_study}

Comparing Table~\ref{table_ablation_study}(b) and \ref{table_ablation_study}(d) shows that proposed monocular SLAM with the scale correction sub-module dramatically improves the trajectory tracking accuracy compared to monocular ORB-SLAM2 without LC. 
Comparing Table \ref{table_ablation_study}(c) and \ref{table_ablation_study}(d) shows that proposed monocular SLAM with the feature refinement sub-module significantly improves the trajectory tracking accuracy compared to monocular ORB-SLAM2 without LC.
In particular, the scale correction sub-module plays an important role in improving proposed monocular SLAM, by resolving the scale ambiguity issue that is a major challenge in monocular SLAM.
The proposed monocular SLAM system can maximize the trajectory tracking accuracy when it uses both the scale correction and the feature refinement sub-modules that require image segmentation process.

\subsection{Comparisons between different visual SLAM methods}
\label{ss_slam_comparison}

\subsubsection{Trajectory tracking accuracy with different visual SLAM methods}

\begin{figure*}[t!]
\centering
\begin{tabular}{ccc}
\includegraphics[width=0.31\linewidth]{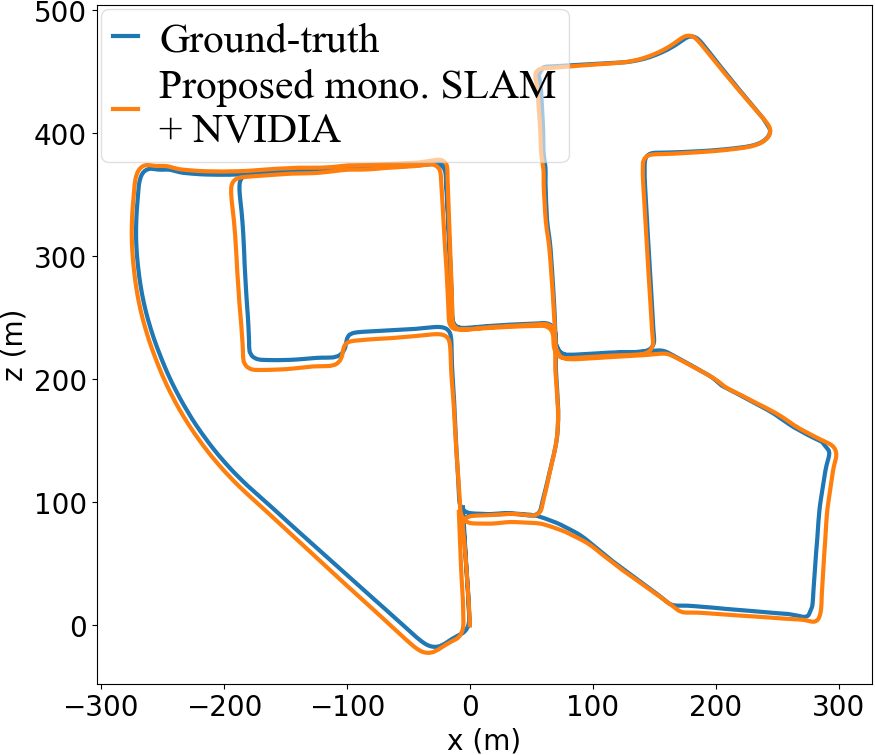} &
\includegraphics[width=0.31\linewidth]{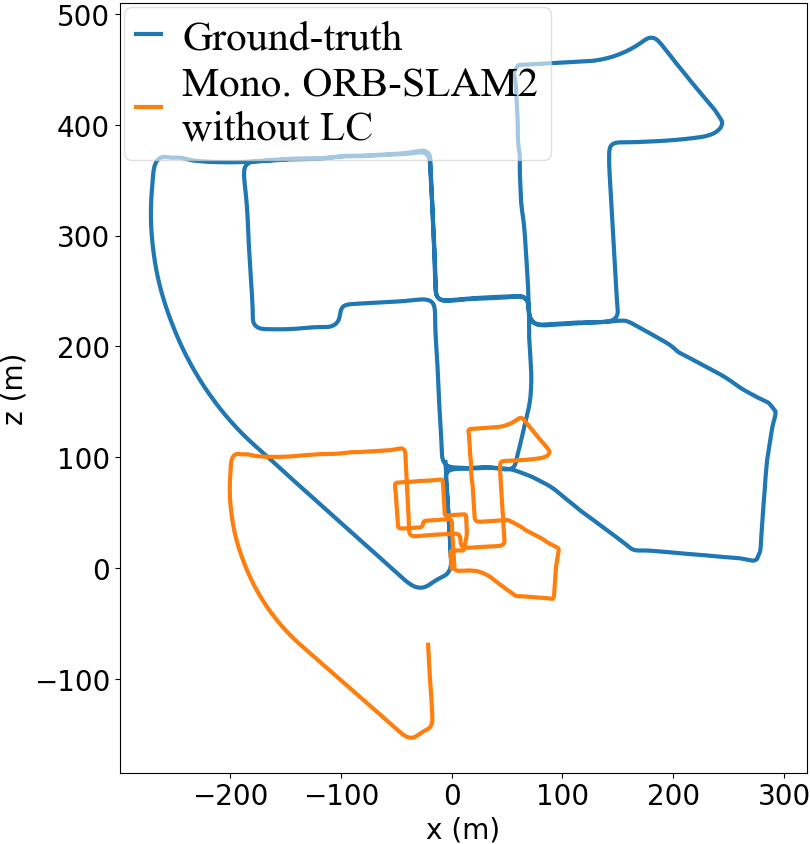} & 
\includegraphics[width=0.31\linewidth]{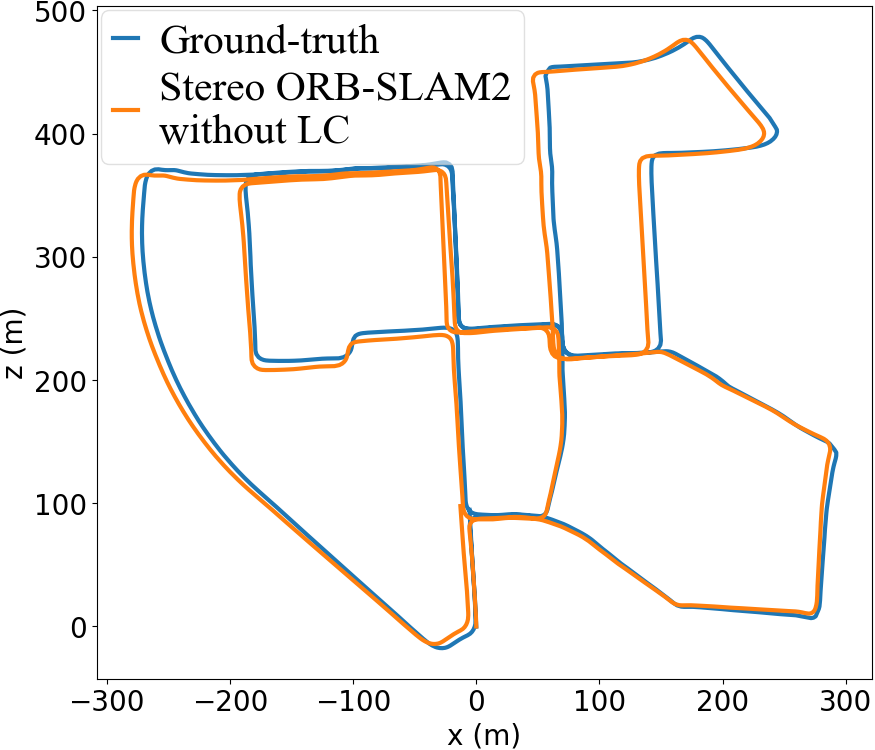} \\
\includegraphics[width=0.31\linewidth]{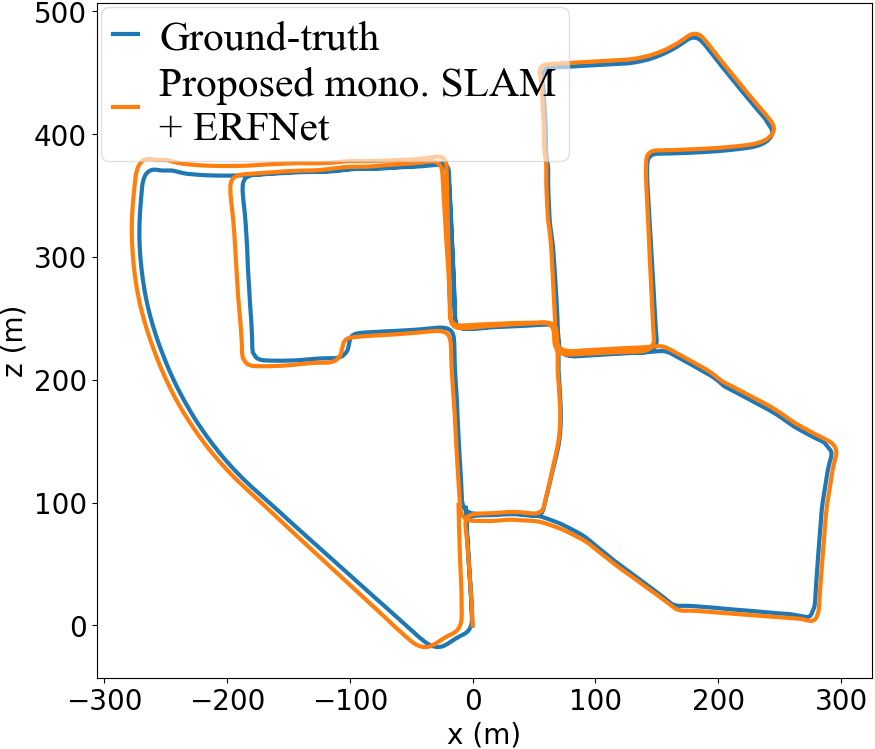} &
\includegraphics[width=0.31\linewidth]{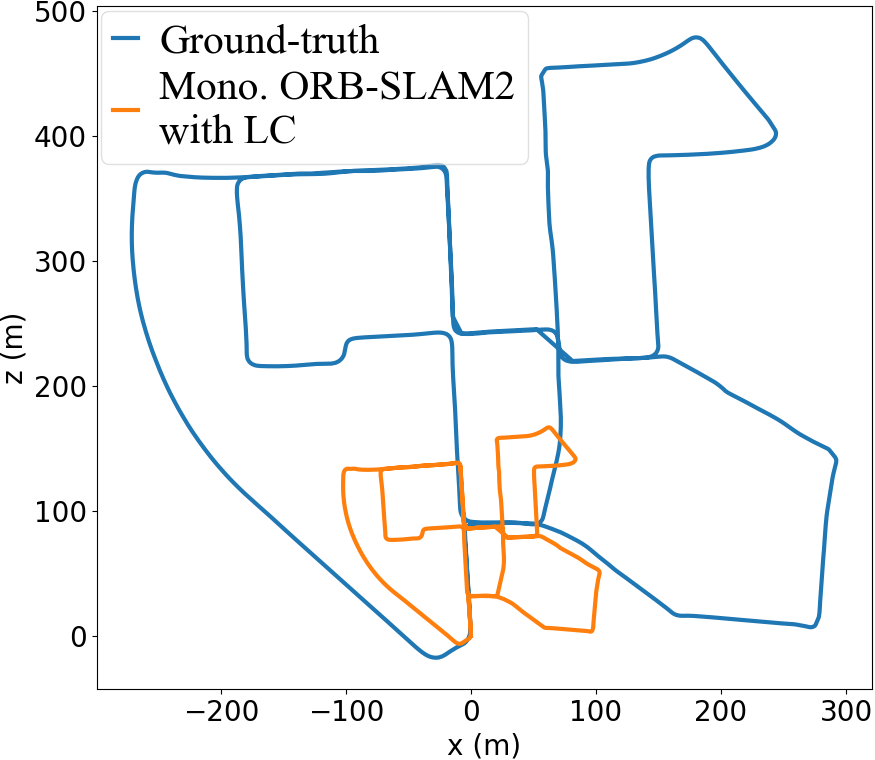} & 
\includegraphics[width=0.31\linewidth]{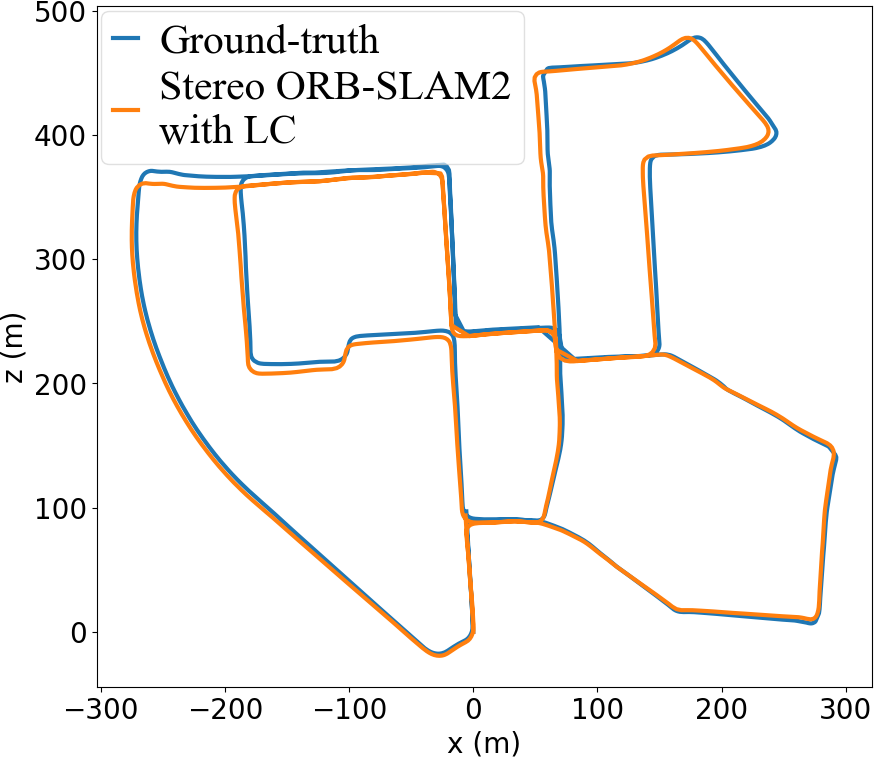}
\end{tabular}
\\
{\footnotesize (a) Sequence $00$ \vspace{0.75pc}}
\\
\begin{tabular}{ccc}
\includegraphics[width=0.31\linewidth]{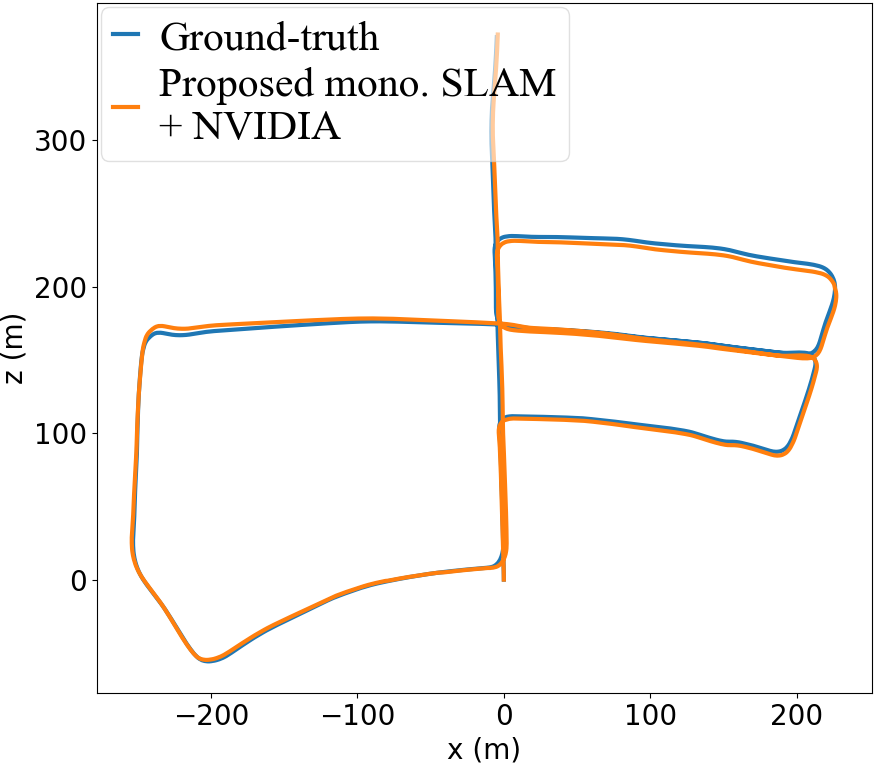} &
\includegraphics[width=0.31\linewidth]{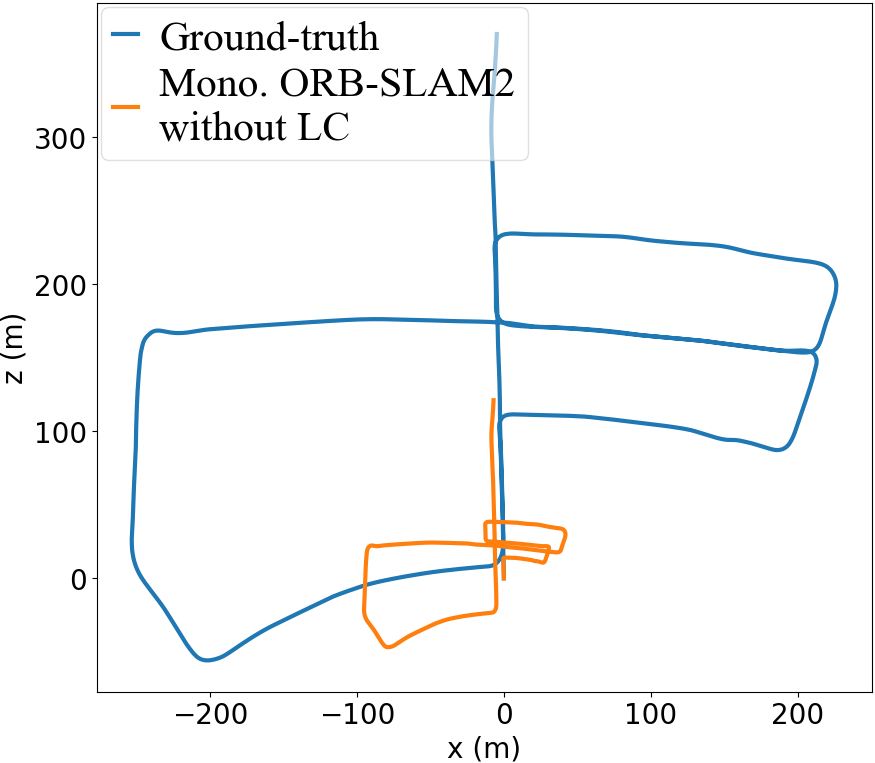} & 
\includegraphics[width=0.31\linewidth]{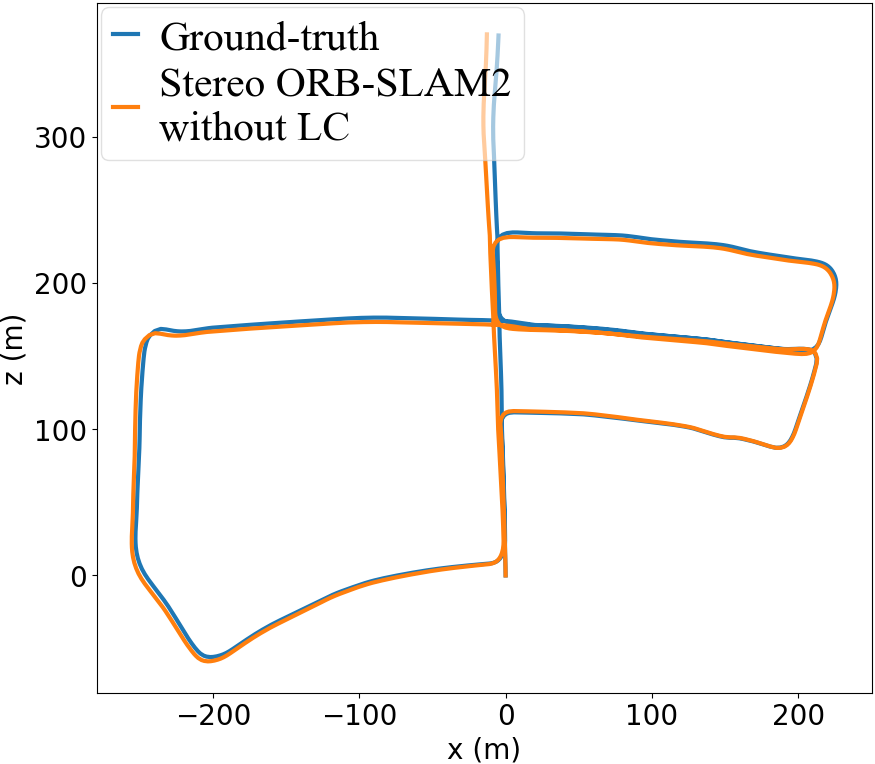} \\
\includegraphics[width=0.31\linewidth]{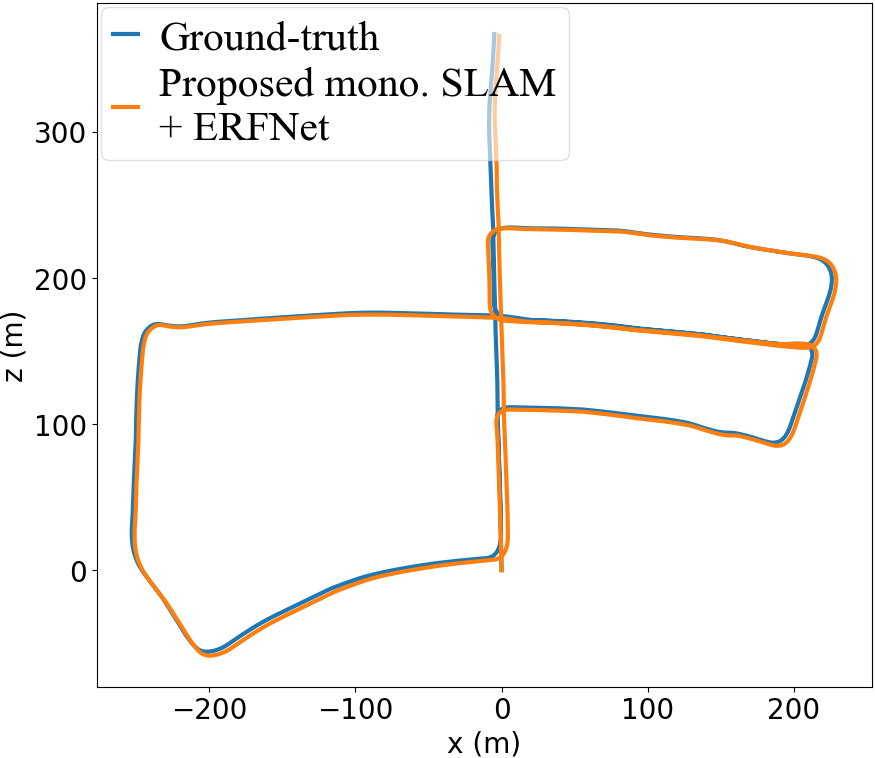} &
\includegraphics[width=0.31\linewidth]{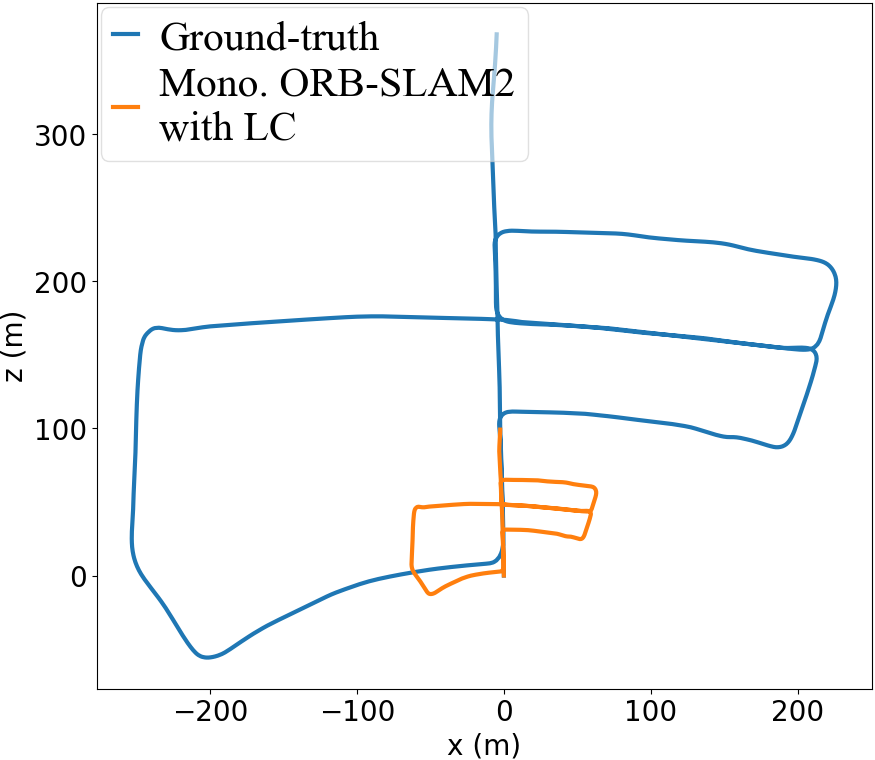} & 
\includegraphics[width=0.31\linewidth]{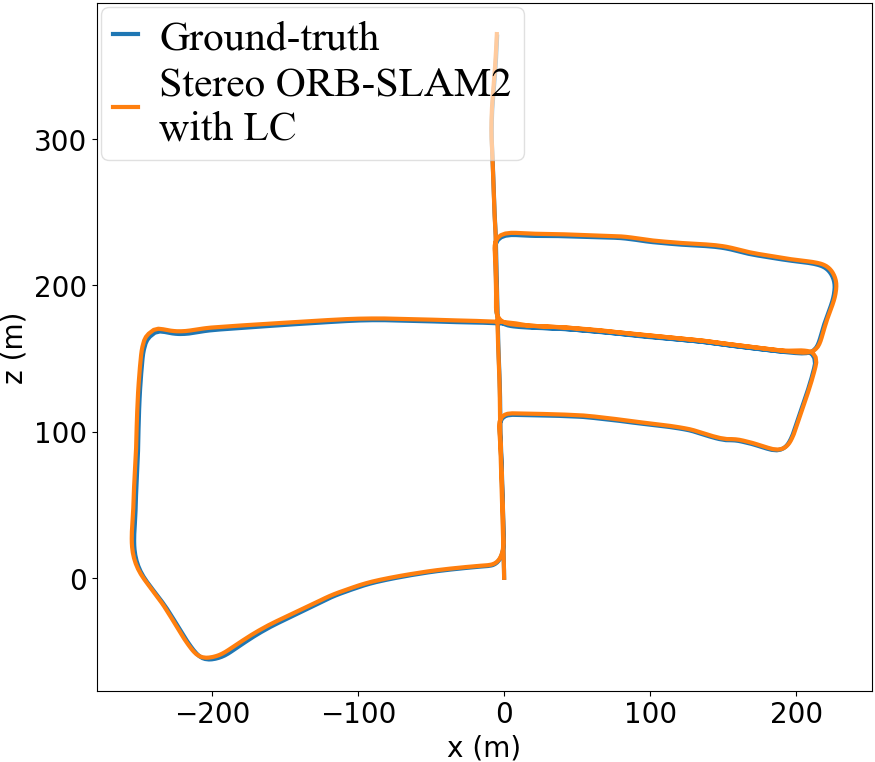}
\end{tabular}
\\
{\footnotesize (b) Sequence $05$
\vspace{-0.25pc}}
\caption{
Qualitative tracking comparisons in the KITTI odometry dataset.
Sequence $00$ and $05$ have a loop or loops in trajectories.
}
\label{fig_qualitative_comparison_loop}
\end{figure*}

\begin{figure*}[t!]
\centering
\begin{tabular}{cc}
\includegraphics[width=0.33\linewidth]{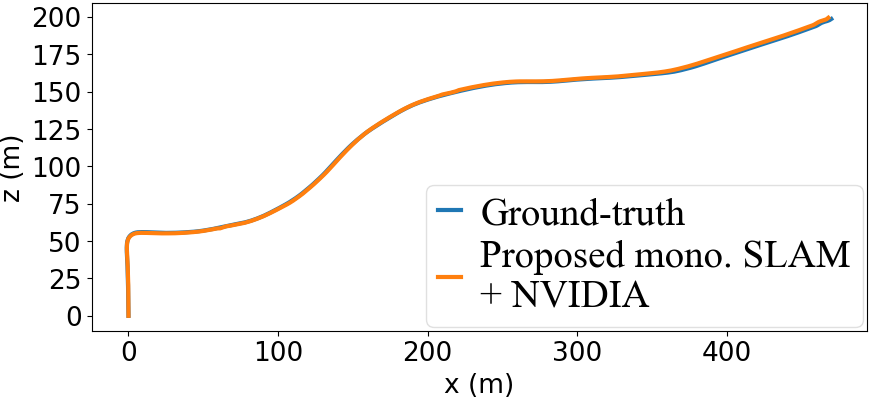} &
\includegraphics[width=0.33\linewidth]{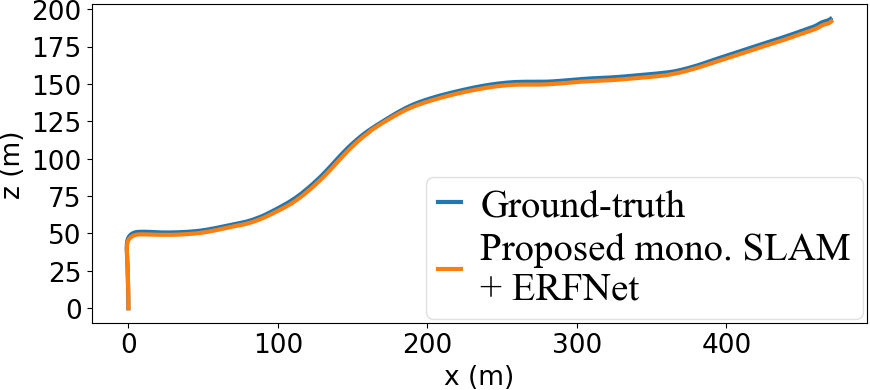} \\
\includegraphics[width=0.33\linewidth]{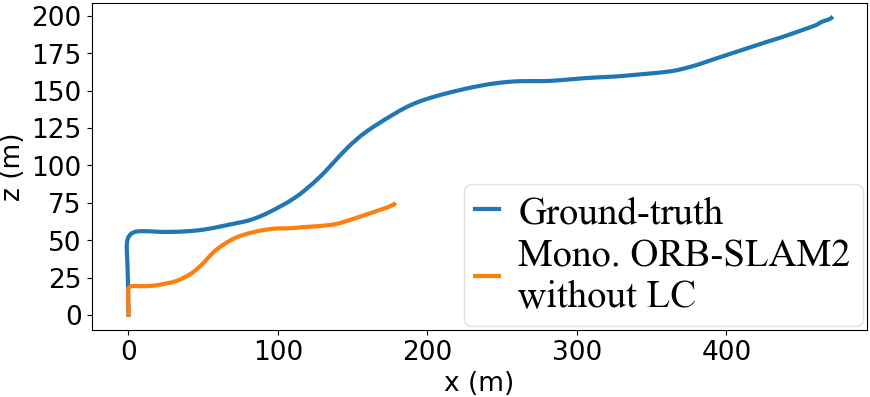} & 
\includegraphics[width=0.33\linewidth]{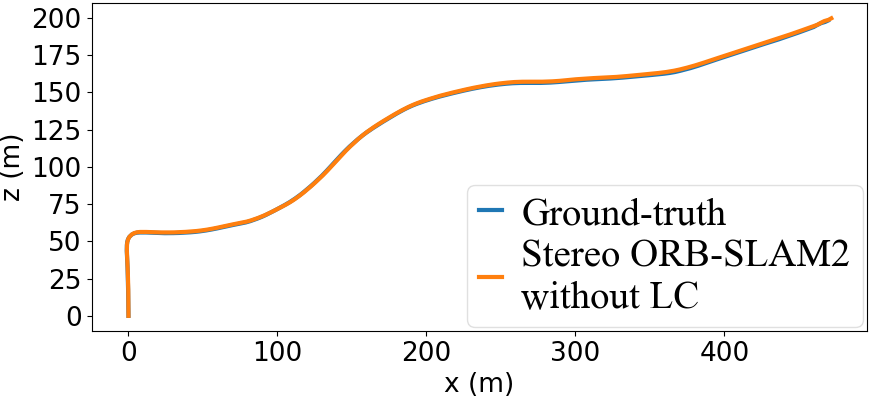}
\end{tabular}
\\
{\footnotesize (a) Sequence $03$ \vspace{0.75pc}}
\\
\begin{tabular}{cc}
\includegraphics[width=0.31\linewidth]{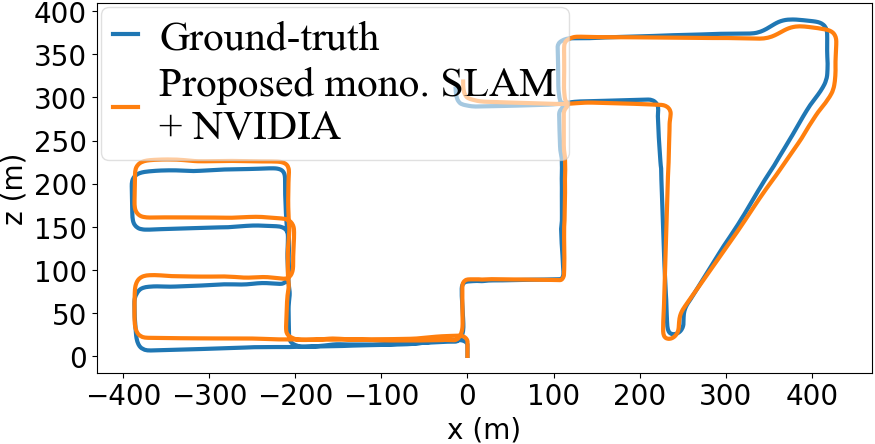} &
\includegraphics[width=0.31\linewidth]{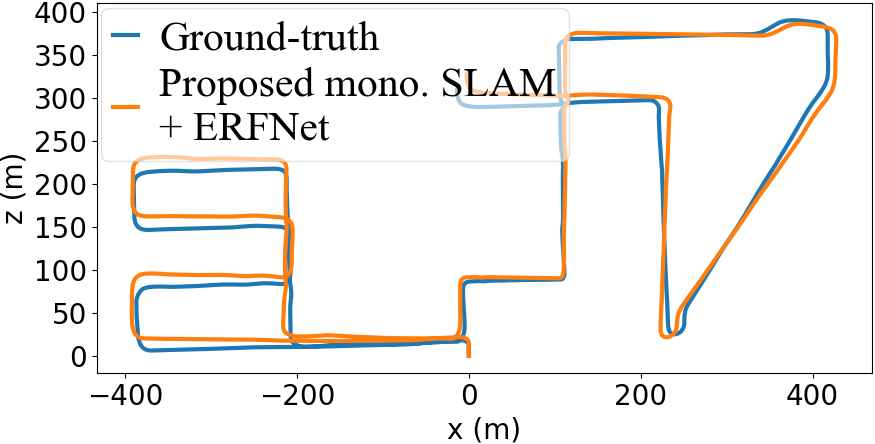} \\
\includegraphics[width=0.31\linewidth]{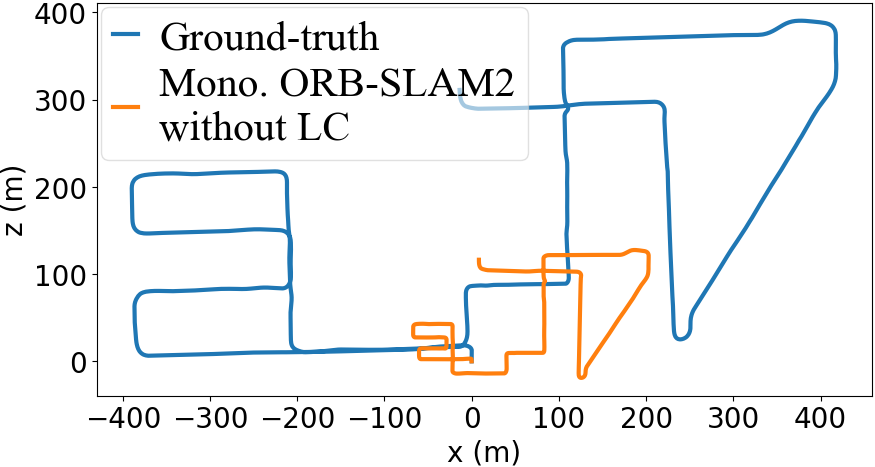} & 
\includegraphics[width=0.31\linewidth]{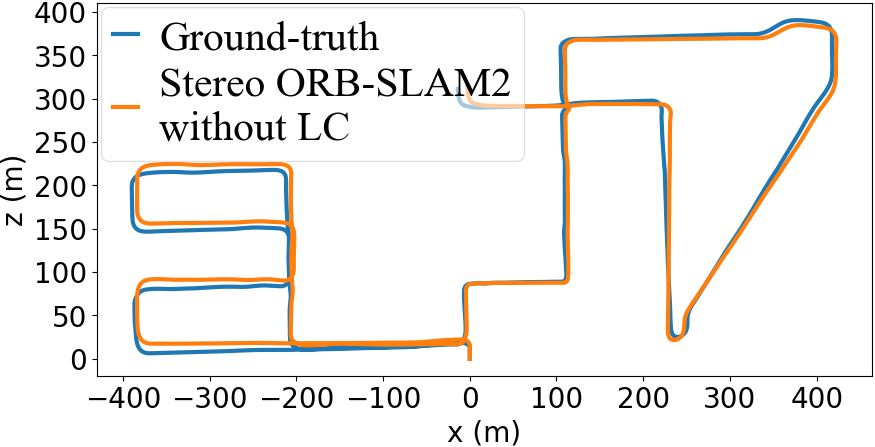}
\end{tabular}
\\
{\footnotesize (b) Sequence $08$
\vspace{-0.25pc}}
\caption{
Qualitative tracking comparisons in the KITTI odometry dataset.
Sequence $03$ and $08$ have no loop(s) in trajectories.
}
\label{fig_qualitative_comparison_no_loop}
\end{figure*}

In all the sequences of the KITTI odometry dataset, regardless of the loop(s) existence,
Tables~\ref{quantitative_eval}(a)--\ref{quantitative_eval}(d) \& \ref{quantitative_eval}(g)--\ref{quantitative_eval}(h) and comparing their trajectories in Figs.~\ref{fig_qualitative_comparison_loop}--\ref{fig_qualitative_comparison_no_loop} show that the proposed monocular SLAM 
with either NVIDIA network on GPU or ERFNet on CPU significantly improves trajectory tracking accuracy compared to monocular ORB-SLAM2 regardless of using LC, and Mask-SLAM.
ERFNet slightly degrades the tracking accuracy of the proposed method, compared to the NVIDIA network.
This is due to the degradation of segmentation accuracy discussed in Section~\ref{exp_decision_resize_factor}.

\begin{figure*}[t!]
\centering
\begin{tabular}{cc}
\includegraphics[width=0.3\linewidth]{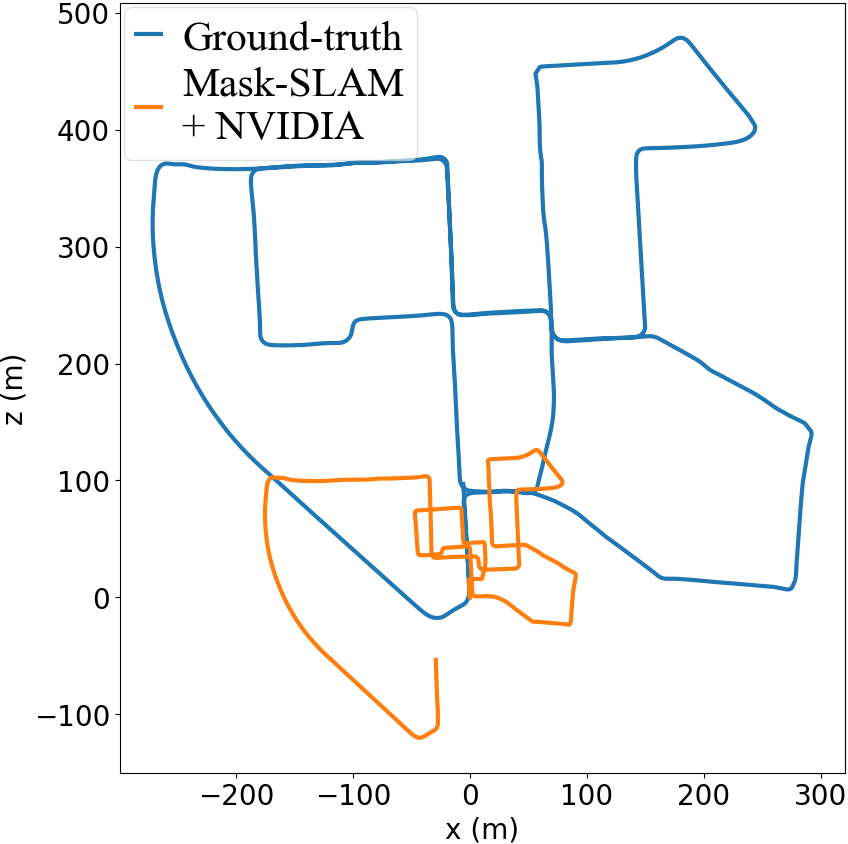} &
\includegraphics[width=0.3\linewidth]{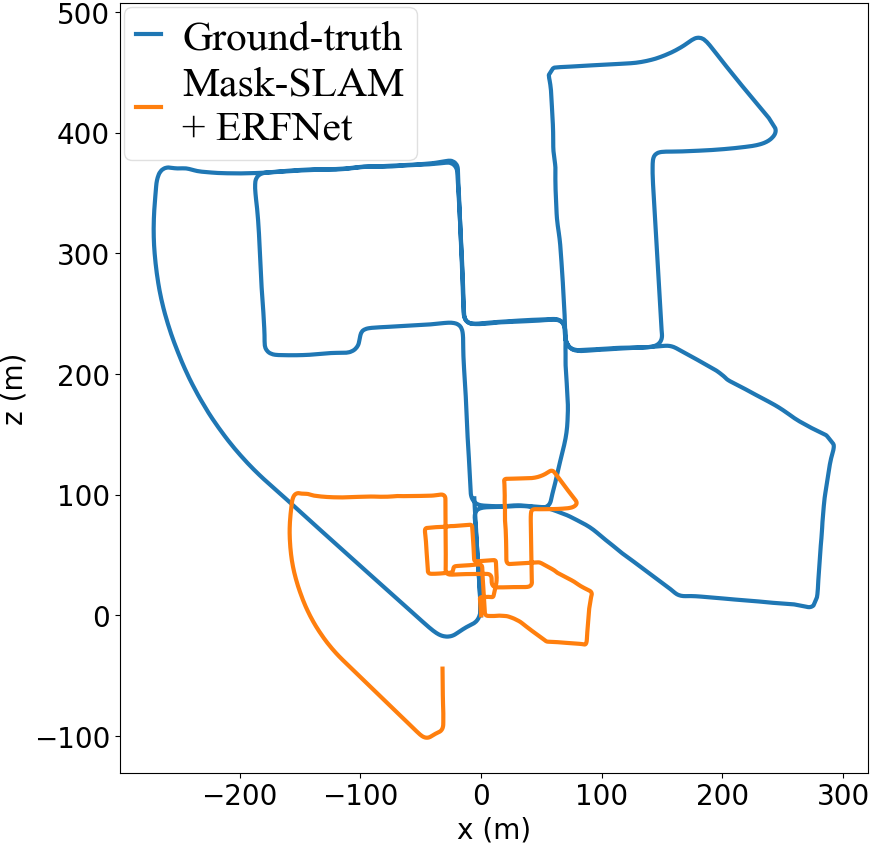}
\end{tabular}
\\
{\footnotesize (a) Sequence $00$ \vspace{0.75pc}}
\\
\begin{tabular}{cc}
\includegraphics[width=0.31\linewidth]{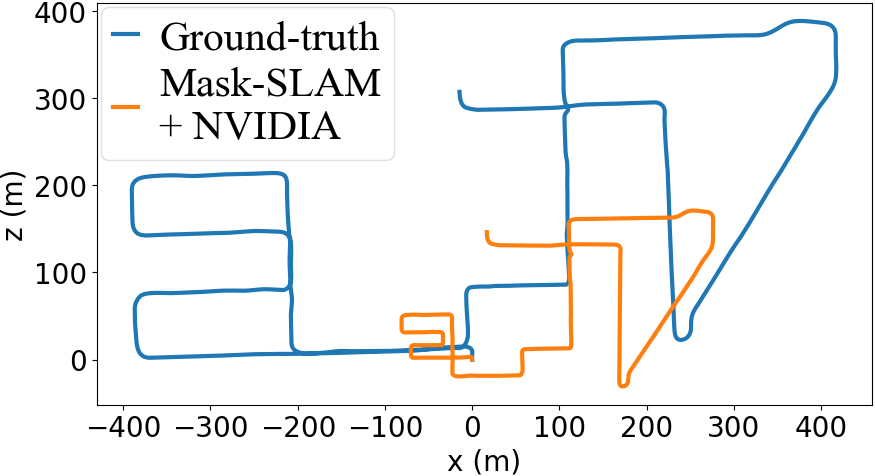} &
\includegraphics[width=0.31\linewidth]{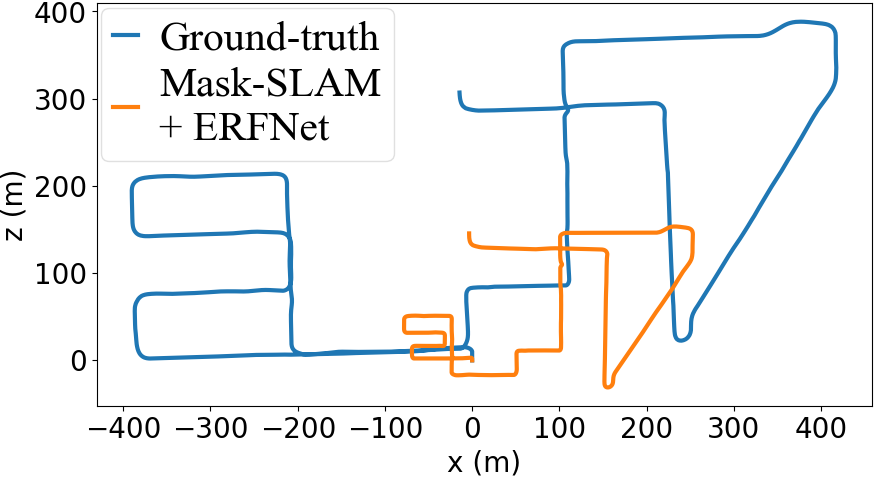}
\end{tabular}
\\
{\footnotesize (b) Sequence $08$
\vspace{-0.25pc}}

\caption{
Qualitative tracking results of Mask-SLAM with different segmentation networks in the KITTI $00$ and $08$ sequences.
}
\label{fig_qualitative_comparison_mask_slam}
\end{figure*}

The proposed monocular visual SLAM system resolves the scale ambiguity by correcting scales using segmented road labels, and achieves significantly better trajectory tracking accuracy.
On the other hand, the monocular ORB-SLAM2 methods suffer from scale ambiguity regardless of LC, their estimated trajectories are off the scale from ground-truth trajectories, and ATE values significantly increase.
Similar to monocular ORB-SLAM2, Mask-SLAM suffers from the scale ambiguity and drift issues, giving poor results in Tables II(g)--II(h) and Fig.~\ref{fig_qualitative_comparison_mask_slam}.

Regardless of the loop(s) existence in the KITTI odometry sequences, 
Tables~\ref{quantitative_eval}(a)--\ref{quantitative_eval}(b), and \ref{quantitative_eval}(e)
and comparing their trajectories in Figs.~\ref{fig_qualitative_comparison_loop}--\ref{fig_qualitative_comparison_no_loop}
demonstrate that 
the trajectory tracking accuracy of the proposed method 
with either GPU-based NVIDIA network or CPU-based ERFNet is comparable to that of stereo ORB-SLAM2 without LC.
The results imply that using only a single camera,
the proposed method can resolve scale ambiguity \& drift and inaccurate mapping issues with comparable performance to stereo ORB-SLAM2 without LC.

In the sequences with loop(s), stereo ORB-SLAM2 with LC gave slightly better trajectory tracking accuracy compared to the proposed method. 
See Tables~\ref{quantitative_eval}(a)--\ref{quantitative_eval}(b) and \ref{quantitative_eval}(f) and compare their results in 
Fig.~\ref{fig_qualitative_comparison_loop}.
In both monocular and stereo ORB-SLAM2s,
the LC scheme improves trajectory tracking accuracy by correcting accumulated localization errors, when a vehicle revisits a previously mapped area (i.e, a loop is created).
Compare results between
Tables~\ref{quantitative_eval}(c)
and \ref{quantitative_eval}(d),
and those between
and Tables~\ref{quantitative_eval}(e)
and \ref{quantitative_eval}(f).
We conjecture by considering the results in Tables~\ref{quantitative_eval}(a)--\ref{quantitative_eval}(b) and Table~\ref{quantitative_eval}(e) that if LC is additionally used,
the proposed monocular SLAM system achieves comparable or better trajectory tracking accuracy, compared to stereo ORB-SLAM2 with LC.

\begin{table*}[t!]
\caption{ATE values (m) of different visual SLAM methods in
different KITTI odometry sequences}
\begin{center}
\begin{tabular}{cc|rrrrrrrr}
\hline\hline
\multicolumn{2}{c|}{Sequence} & \multicolumn{1}{c}{\begin{tabular}[c]{@{}c@{}}(a) Proposed\\ mono.~SLAM\\ + NVIDIA$^\dagger$\end{tabular}} & \multicolumn{1}{c}{\begin{tabular}[c]{@{}c@{}}(b) Proposed\\ mono.~SLAM\\ + ERFNet$^\ddagger$ \end{tabular}} & \multicolumn{1}{c}{\begin{tabular}[c]{@{}c@{}}(c) Monocular\\ ORB-SLAM2\\ without LC\end{tabular}} & \multicolumn{1}{c}{\begin{tabular}[c]{@{}c@{}}(d) Monocular\\ ORB-SLAM2\\ with LC\end{tabular}} & \multicolumn{1}{c}{\begin{tabular}[c]{@{}c@{}}(e) Stereo\\ ORB-SLAM2\\ without LC\end{tabular}} & \multicolumn{1}{c}{\begin{tabular}[c]{@{}c@{}}(f) Stereo\\ ORB-SLAM2\\ with LC\end{tabular}} &
\multicolumn{1}{c}{\begin{tabular}[c]{@{}c@{}}(g) Mask-\\ SLAM\\ + NVIDIA$^\dagger$\end{tabular}} &
\multicolumn{1}{c}{\begin{tabular}[c]{@{}c@{}}(h) Mask-\\ SLAM\\ + ERFNet$^\ddagger$\end{tabular}} \\ \hline
\multicolumn{1}{c|}{\multirow{4}{*}{\begin{tabular}[c]{@{}c@{}}With\\ loop(s)\end{tabular}}} & 00 & \textbf{6.797} & 8.451 & 231.701 & 199.658 & 8.515 & 6.869 & 229.908 & 231.473\\
\multicolumn{1}{c|}{} & 02 & \textbf{13.059} & 14.874 & 550.845 & 421.711 & 14.330 & 13.437 & 523.694 & 516.256\\
\multicolumn{1}{c|}{} & 05 & 2.888 & 3.763 & 170.815 & 143.368 & 2.949 & \textbf{1.567} & 153.925 & 154.101\\
\multicolumn{1}{c|}{} & 07 & 2.354 & 3.464 & 80.490 & 67.158 & 2.856 & \textbf{1.057} & 71.778 & 72.629\\ \hline
\multicolumn{1}{c|}{\multirow{3}{*}{\begin{tabular}[c]{@{}c@{}}No\\ loops\end{tabular}}} & 03 & \textbf{1.306} & 2.087 & 140.736 & - & 1.387 & - & 191.354 & 193.940\\
\multicolumn{1}{c|}{} & 04 & 1.271 & 1.730 & 200.135 & - & \textbf{1.008} & - & 191.348 & 191.578\\
\multicolumn{1}{c|}{} & 08 & 12.973 & 13.428 & 196.803 & - & \textbf{12.765} & - & 195.471 & 203.622
\\ \hline\hline
\end{tabular}
\smallskip
\begin{myquote}{0.3in}
$^\dagger$ The NVIDIA semantic-segmentation network \cite{zhu2019improving} ran on a standard GPU with keyframes downsampled by factor 2.\\
$^\ddagger$ ERFNet segmentation \cite{romera2017erfnet} ran on a standard CPU with non-downsampled keyframes.
\end{myquote}
\end{center}
\label{quantitative_eval}
\end{table*}

\begin{table}[t!]
\centering
\caption{
Tracking time per frame with different monocular SLAM methods using segmentation (KITTI sequences in Table~\ref{quantitative_eval})
}
\begin{tabular}{l|rr}
\hline\hline
\multicolumn{1}{c|}{\multirow{2}{*}{\begin{tabular}[c]{@{}c@{}}Monocular SLAM methods \\ using segmentation\end{tabular}}} & \multicolumn{2}{c}{\begin{tabular}[c]{@{}c@{}}Tracking time\\ per frame (ms)\end{tabular}} \\ \cline{2-3} 
\multicolumn{1}{c|}{} & \multicolumn{1}{c}{Mean} & \multicolumn{1}{c}{Std.} \\ \hline
Proposed mono.~SLAM + NVIDIA$^\dagger$ & 28.52 & 7.77 \\
Proposed mono.~SLAM + ERFNet$^\ddagger$ & 29.74 & 9.95 \\ \hline
Mask-SLAM + NVIDIA$^\dagger$ & 145.51 & 9.96 \\
Mask-SLAM + ERFNet$^\ddagger$ & 270.63 & 9.30 \\ 
\hline\hline
\end{tabular}
\smallskip
\begin{myquote}{0.3in}
$^\dagger$ The NVIDIA semantic-segmentation network \cite{zhu2019improving} ran on a standard GPU with keyframes downsampled by factor 2.\\
$^\ddagger$ ERFNet segmentation \cite{romera2017erfnet} ran on a standard CPU with non-downsampled keyframes.
\end{myquote}
\label{fps_comparison}
\end{table}

\begin{table}[t!]
\centering
\caption{Processing time for each (sub-)modules (KITTI-00 dataset)}
\begin{tabular}{c|l|cc}
\hline\hline
\multirow{2}{*}{Modules} & \multicolumn{1}{c|}{\multirow{2}{*}{Sub-modules}} & \multicolumn{2}{c}{\begin{tabular}[c]{@{}c@{}} Processing time \\ (ms)\end{tabular}} \\ \cline{3-4} 
 & \multicolumn{1}{c|}{} & Mean & Std. \\ \hline
 \begin{tabular}[c]{@{}c@{}} Localization \\ (per frame) \end{tabular} & Total & ~27.87 & ~7.93 \\ \hline
\multirow{3}{*}{\begin{tabular}[c]{@{}c@{}} Mapping \\ (per keyframe) \end{tabular}} & Triangulation & 130.10 & 43.01 \\
 & Scale correction & ~14.99 & 32.45 \\ \cline{2-4} 
 & Total & 146.55 & 73.64 \\ \hline
\multirow{3}{*}{\begin{tabular}[c]{@{}c@{}} Segmentation \\ (per keyframe) \end{tabular}} & Semantic segmentation $^\dagger$ & 129.44 & 4.13 \\
 & Feature refinement  & ~70.31 & 30.60 \\ \cline{2-4}
 & Total & 201.74 & 31.27 \\ 
\hline\hline
\end{tabular}
\smallskip
\begin{myquote}{0.3in}$^\dagger$ The NVIDIA semantic-segmentation network \cite{zhu2019improving} ran on a standard GPU with keyframes downsampled by factor 2.
\end{myquote}
\label{modules_processing_time}
\end{table}

\begin{figure}[t!]
\centering
\includegraphics[width=\linewidth]{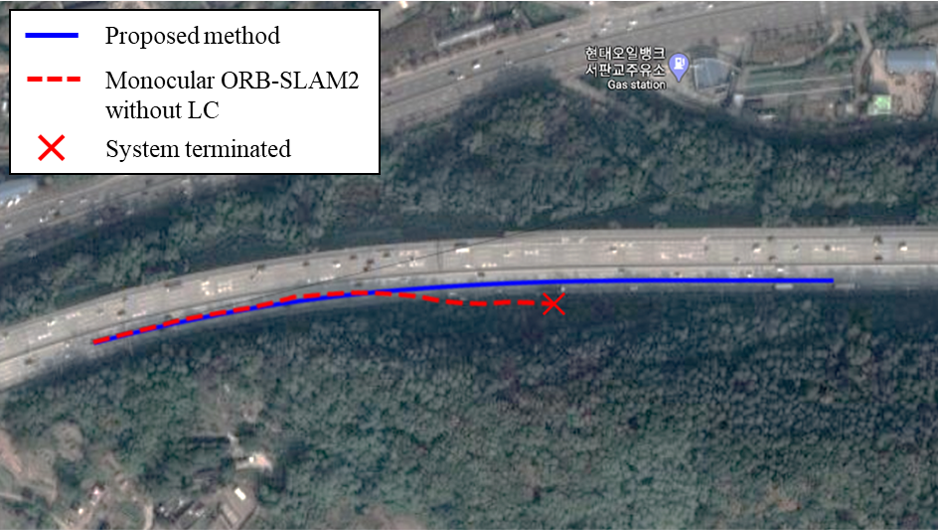}
\caption{Comparison of the proposed method (with the NVIDIA network on a standard GPU), and monocular ORB-SLAM2 without LC in the StradVision sequence.
The blue solid and red dashed lines denote 
estimated trajectory by the proposed method and
monocular ORB-SLAM2 without LC, respectively.
The estimated trajectories are aligned with the map in Google Maps.
We observed a very similar tracking performance between the proposed systems using the GPU-based NVIDIA net and and CPU-based ERFNet.}
\label{fig_experiment_3}
\end{figure}

In the StradVision sequence (that has no loops), 
Fig.~\ref{fig_experiment_3} shows that the proposed method significantly improves the trajectory tracking accuracy compared to monocular ORB-SLAM2 without LC.
Whereas the proposed method consistently followed a road route,
monocular ORB-SLAM2 without LC gave deviated vehicle localization at the $535$th frame,
and was eventually terminated at the $749$th frame.
The monocular ORB-SLAM2 system using no LC does not remove 
moving objects and low-parallax areas in generating 3D points at its mapping module.
Such inappropriate 3D points could significantly degrade camera pose estimation performance, and eventually may terminate the entire system.
The proposed SLAM system overcomes that challenge by corner feature refinements in the segmentation thread; see Section~\ref{ss_incorrect_mapping_reduction}.

\subsubsection{Comparisons of real-time execution feasibility of different monocular SLAMs using semantic segmentation}
\label{experiment_comparison_real_time}

Table \ref{fps_comparison} compares tracking time per frame between the proposed monocular SLAM system and Mask-SLAM, given either NVIDIA network on a standard GPU or ERFNet on a standard CPU
(see CPU and GPU specs in Section~\ref{ss_experimental_setup}).
Regardless of segmentation methods,
the averaged tracking time per frame of the proposed system is below the inverse of the camera frame-rate, 100 ms, implying that the system can run in real-time \cite{mur2017orb} on a standard CPU and/or GPU.
The proposed efficient integration of semantic segmentation with monocular SLAM achieves real-time execution by \textit{1)} applying semantic segmentation to keyframes and \textit{2)} sophisticatedly considering localization and mapping timings; 
see details in Section~\ref{ss_efficient_integration}.
Since the averaged tracking time per frame of the proposed method is below 33 ms given imaging resolution $1241\times376$, we conjecture that the proposed method runs in real-time with some $30$ Hz cameras 
with an imaging resolution lower than or equal to $1241\times376$.
Different from the proposed method, 
Table \ref{fps_comparison} implies that Mask-SLAM is infeasible for real-time running on a standard CPU or GPU. 
This aligns with the claim in \cite[Sec. 5]{kaneko2018mask}.

The proposed monocular SLAM
-- with either GPU-based segmentation using the NVIDIA network and the keyframe downsampling factor 2 or CPU-based segmentation using ERFNet --
did not experience system termination/failure in all the eight sequences.
One might operate the proposed monocular SLAM system with the NVIDIA segmentation network for high-quality tracking in real-time, using some GPU-based vehicle board, e.g., NVIDIA Jetson \cite{jetson}.
If a standard GPU is unavailable, the proposed system with EPFNet may run in real-time on a standard CPU, but at little expense of tracking accuracy.

We note that the processing time of the localization module determines the real-time execution feasibility of the proposed method, because all modules run in different threads in parallel and the localization module is the only module that runs for every frames.
In Table~\ref{modules_processing_time}, the averaged processing times of the mapping and the segmentation modules are greater than 100~ms, but they run in parallel only for every \textit{keyframe}.

\section{Conclusion}
\label{sec:conclusion}

Monocular SLAM methods use only a single camera and thus, have major limitations in scale estimation and mapping accuracy.
The proposed monocular SLAM system resolves the limitations by efficiently using deep learning-based semantic segmentation: 
it applies a state-of-the-art semantic segmentation only to (downsampled) keyframes in parallel with mapping processes.
The proposed method performs scale correction using road-labeled 3D points and removes features labeled by dynamic objects and low parallax areas in 3D point generations.
Regardless of the existence of loops in video sequences, proposed monocular SLAM achieves significantly better trajectory tracking accuracy compared to state-of-the-art monocular SLAM, monocular ORB-SLAM2 \cite{mur2017orb}.
In video sequences without loops, proposed monocular SLAM achieves comparable trajectory tracking accuracy compared to state-of-the-art stereo SLAM, stereo ORB-SLAM2 \cite{mur2017orb}.
Compared with an existing segmentation-aided monocular SLAM method, Mask-SLAM \cite{kaneko2018mask},
the proposed system achieves significantly improved trajectory tracking accuracy and reduced tracking time per frame; it can run in real time on a standard CPU potentially with a standard GPU.
We believe that the proposed monocular SLAM system would be particularly useful in ADAS and autonomous driving systems, by using only a single camera but having comparable performances to stereo visual SLAM systems.
The proposed efficient integration of semantic segmentation with monocular SLAM could be applicable to a visual SLAM integration with other deep-learning methods needing high computation costs, such as depth estimation.

There are a number of avenues for future work.
First, we will additionally run LC in an additional thread, 
to further improve trajectory tracking accuracy.
Second, we additionally incorporate a relocalization technique (e.g., \cite{eade2008unified,williams2011automatic,mur2014fast}) into the proposed method to improve the robustness against tracking failure.
Third, we plan to use camera poses, 3D point coordinates, etc.~to refine segmentation results, motivated by \cite{wang2019unified}.
On the image sensor side, the future work is modifying the proposed framework for fish-eye or omnidirectional camera.

\bibliographystyle{IEEEtran}
\bibliography{IEEEabrv}

\begin{thebibliography}{10}
\providecommand{\url}[1]{#1}
\csname url@samestyle\endcsname
\providecommand{\newblock}{\relax}
\providecommand{\bibinfo}[2]{#2}
\providecommand{\BIBentrySTDinterwordspacing}{\spaceskip=0pt\relax}
\providecommand{\BIBentryALTinterwordstretchfactor}{4}
\providecommand{\BIBentryALTinterwordspacing}{\spaceskip=\fontdimen2\font plus
\BIBentryALTinterwordstretchfactor\fontdimen3\font minus
  \fontdimen4\font\relax}
\providecommand{\BIBforeignlanguage}[2]{{%
\expandafter\ifx\csname l@#1\endcsname\relax
\typeout{** WARNING: IEEEtran.bst: No hyphenation pattern has been}%
\typeout{** loaded for the language `#1'. Using the pattern for}%
\typeout{** the default language instead.}%
\else
\language=\csname l@#1\endcsname
\fi
#2}}
\providecommand{\BIBdecl}{\relax}
\BIBdecl

\bibitem{davison2007monoslam}
A.~J. Davison, I.~D. Reid, N.~D. Molton, and O.~Stasse, ``Mono{SLAM}: Real-time
  single camera {SLAM},'' \emph{IEEE Transactions on Pattern Analysis and
  Machine Intelligence}, vol.~29, no.~6, pp. 1052--1067, June 2007.

\bibitem{civera2008inverse}
J.~Civera, A.~J. Davison, and J.~M. Montiel, ``Inverse depth parametrization
  for monocular {SLAM},'' \emph{IEEE Transactions on Robotics}, vol.~24, no.~5,
  pp. 932--945, October 2008.

\bibitem{chiuso2002structure}
A.~Chiuso, P.~Favaro, H.~Jin, and S.~Soatto, ``Structure from motion causally
  integrated over time,'' \emph{IEEE Transactions on Pattern Analysis and
  Machine Intelligence}, vol.~24, no.~4, pp. 523--535, April 2002.

\bibitem{eade2006scalable}
E.~Eade and T.~Drummond, ``Scalable monocular {SLAM},'' in \emph{Proc.~IEEE
  Conference on Computer Vision and Pattern Recognition}, New York, NY, June
  2006, pp. 469--476.

\bibitem{younes2017keyframe}
G.~Younes, D.~Asmar, E.~Shammas, and J.~Zelek, ``Keyframe-based monocular
  {SLAM}: design, survey, and future directions,'' \emph{Robotics and
  Autonomous Systems}, vol.~98, pp. 67--88, December 2017.

\bibitem{mouragnon2006real}
E.~Mouragnon, M.~Lhuillier, M.~Dhome, F.~Dekeyser, and P.~Sayd, ``Real time
  localization and 3d reconstruction,'' in \emph{Proc.~IEEE Conference on
  Computer Vision and Pattern Recognition}, vol.~1, New York, NY, June 2006,
  pp. 363--370.

\bibitem{klein2007parallel}
G.~Klein and D.~Murray, ``Parallel tracking and mapping for small ar
  workspaces,'' in \emph{Proc.~IEEE and ACM International Symposium on Mixed
  and Augmented Reality}, Nara, Japan, November 2007, pp. 225--234.

\bibitem{mur2015orb}
R.~Mur-Artal, J.~M.~M. Montiel, and J.~D. Tardos, ``{ORB-SLAM}: {A} versatile
  and accurate monocular {SLAM} system,'' \emph{IEEE Transactions on Robotics},
  vol.~31, no.~5, pp. 1147--1163, August 2015.

\bibitem{strasdat2012visual}
H.~Strasdat, J.~M. Montiel, and A.~J. Davison, ``Visual {SLAM}: why filter?''
  \emph{Image and Vision Computing}, vol.~30, no.~2, pp. 65--77, February 2012.

\bibitem{zhou2019ground}
D.~Zhou, Y.~Dai, and H.~Li, ``Ground plane based absolute scale estimation for
  monocular visual odometry,'' \emph{IEEE Transactions on Intelligent
  Transportation Systems}, vol.~21, no.~2, pp. 791--802, May 2019.

\bibitem{hartley2003multiple}
R.~Hartley and A.~Zisserman, \emph{Multiple View Geometry in Computer Vision},
  2nd~ed.\hskip 1em plus 0.5em minus 0.4em\relax Cambridge University Press,
  March 2004.

\bibitem{grater2015robust}
J.~Gr{\"a}ter, T.~Schwarze, and M.~Lauer, ``Robust scale estimation for
  monocular visual odometry using structure from motion and vanishing points,''
  in \emph{Proc.~IEEE Intelligent Vehicles Symposium}, Seoul, Korea, June 2015,
  pp. 475--480.

\bibitem{lovegrove2011accurate}
S.~Lovegrove, A.~J. Davison, and J.~Ibanez-Guzm{\'a}n, ``Accurate visual
  odometry from a rear parking camera,'' in \emph{Proc.~IEEE Intelligent
  Vehicles Symposium}, Baden-Baden, Germany, June 2011, pp. 788--793.

\bibitem{mirabdollah2015fast}
M.~H. Mirabdollah and B.~Mertsching, ``Fast techniques for monocular visual
  odometry,'' in \emph{Proc.~German Conference on Pattern Recognition}, Aachen,
  Germany, October 2015, pp. 297--307.

\bibitem{pereira2017monocular}
F.~I. Pereira, G.~Ilha, J.~Luft, M.~Negreiros, and A.~Susin, ``Monocular visual
  odometry with cyclic estimation,'' in \emph{Proc.~SIBGRAPI Conference on
  Graphics, Patterns and Images}, Niteroi, Brazil, October 2017, pp. 1--6.

\bibitem{scaramuzza2009absolute}
D.~Scaramuzza, F.~Fraundorfer, M.~Pollefeys, and R.~Siegwart, ``Absolute scale
  in structure from motion from a single vehicle mounted camera by exploiting
  nonholonomic constraints,'' in \emph{Proc.~IEEE International Conference on
  Computer Vision}, Kyoto, Japan, September 2009, pp. 1413--1419.

\bibitem{yang2019cubeslam}
S.~Yang and S.~Scherer, ``Cube{SLAM}: Monocular 3-{D} object {SLAM},''
  \emph{IEEE Transactions on Robotics}, vol.~35, no.~4, pp. 925--938, August
  2019.

\bibitem{botterill2011bag}
T.~Botterill, S.~Mills, and R.~Green, ``Bag-of-words-driven, single-camera
  simultaneous localization and mapping,'' \emph{Journal of Field Robotics},
  vol.~28, no.~2, pp. 204--226, March 2011.

\bibitem{botterill2012correcting}
------, ``Correcting scale drift by object recognition in single-camera
  {SLAM},'' \emph{IEEE Transactions on Cybernetics}, vol.~43, no.~6, pp.
  1767--1780, December 2012.

\bibitem{song2015high}
S.~Song, M.~Chandraker, and C.~C. Guest, ``High accuracy monocular {SFM} and
  scale correction for autonomous driving,'' \emph{IEEE Transactions on Pattern
  Analysis and Machine Intelligence}, vol.~38, no.~4, pp. 730--743, April 2016.

\bibitem{strasdat2010scale}
H.~Strasdat, J.~Montiel, and A.~J. Davison, ``Scale drift-aware large scale
  monocular {SLAM},'' \emph{Robotics: Science and Systems}, vol.~2, no.~3,
  p.~7, June 2010.

\bibitem{an2017semantic}
L.~An, X.~Zhang, H.~Gao, and Y.~Liu, ``Semantic segmentation-aided visual
  odometry for urban autonomous driving,'' \emph{International Journal of
  Advanced Robotic Systems}, vol.~14, no.~5, pp. 1--11, October 2017.

\bibitem{brasch2018semantic}
N.~Brasch, A.~Bozic, J.~Lallemand, and F.~Tombari, ``Semantic monocular {SLAM}
  for highly dynamic environments,'' in \emph{Proc.~IEEE/RSJ International
  Conference on Intelligent Robots and Systems}, Madrid, Spain, October 2018,
  pp. 393--400.

\bibitem{kaneko2018mask}
M.~Kaneko, K.~Iwami, T.~Ogawa, T.~Yamasaki, and K.~Aizawa, ``Mask-{SLAM}:
  Robust feature-based monocular {SLAM} by masking using semantic
  segmentation,'' in \emph{Proc.~IEEE Conference on Computer Vision and Pattern
  Recognition Workshops}, Salt Lake City, UT, June 2018, pp. 258--266.

\bibitem{yu2018ds}
C.~Yu, Z.~Liu, X.-J. Liu, F.~Xie, Y.~Yang, Q.~Wei, and Q.~Fei, ``{DS-SLAM}: A
  semantic visual {SLAM} towards dynamic environments,'' in
  \emph{Proc.~IEEE/RSJ International Conference on Intelligent Robots and
  Systems}, Madrid, Spain, October 2018, pp. 1168--1174.

\bibitem{baker2004lucas}
S.~Baker and I.~Matthews, ``Lucas-{K}anade 20 years on: A unifying framework,''
  \emph{International journal of computer vision}, vol.~56, no.~3, pp.
  221--255, February 2004.

\bibitem{rublee2011orb}
E.~Rublee, V.~Rabaud, K.~Konolige, and G.~Bradski, ``{ORB}: An efficient
  alternative to {SIFT} or {SURF},'' in \emph{Proc.~IEEE International
  Conference on Computer Vision}, Barcelona, Spain, November 2011, pp.
  2564--2571.

\bibitem{anton2013elementary}
H.~Anton and C.~Rorres, \emph{Elementary Linear Algebra}.\hskip 1em plus 0.5em
  minus 0.4em\relax John Wiley \& Sons, September 2013.

\bibitem{fischler1981random}
M.~A. Fischler and R.~C. Bolles, ``Random sample consensus: a paradigm for
  model fitting with applications to image analysis and automated
  cartography,'' \emph{Communications of the ACM}, vol.~24, no.~6, pp.
  381--395, June 1981.

\bibitem{mur2017orb}
R.~Mur-Artal and J.~D. Tard{\'o}s, ``{ORB}-{SLAM}2: An open-source {SLAM}
  system for monocular, stereo, and {RGB-D} cameras,'' \emph{IEEE Transactions
  on Robotics}, vol.~33, no.~5, pp. 1255--1262, June 2017.

\bibitem{scaramuzza2011visual}
D.~Scaramuzza and F.~Fraundorfer, ``Visual odometry: Part 1: The first 30 years
  and fundamentals,'' \emph{IEEE robotics \& automation magazine}, vol.~18,
  no.~4, pp. 80--92, December 2011.

\bibitem{eade2008unified}
E.~Eade and T.~Drummond, ``Unified loop closing and recovery for real time
  monocular {SLAM}.'' in \emph{Proc.~British Machine Vision Conference},
  vol.~13, Leeds, UK, June 2008, p. 136.

\bibitem{williams2011automatic}
B.~Williams, G.~Klein, and I.~Reid, ``Automatic relocalization and loop closing
  for real-time monocular {SLAM},'' \emph{IEEE Transactions on Pattern Analysis
  and Machine Intelligence}, vol.~33, no.~9, pp. 1699--1712, September 2011.

\bibitem{mur2014fast}
R.~Mur-Artal and J.~D. Tard{\'o}s, ``Fast relocalisation and loop closing in
  keyframe-based {SLAM},'' in \emph{Proc.~IEEE International Conference on
  Robotics and Automation (ICRA)}, Hong Kong, China, May 2014, pp. 846--853.

\bibitem{sturm2012benchmark}
J.~Sturm, N.~Engelhard, F.~Endres, W.~Burgard, and D.~Cremers, ``A benchmark
  for the evaluation of {RGB-D SLAM} systems,'' in \emph{Proc.~IEEE/RSJ
  International Conference on Intelligent Robots and Systems},
  Vilamoura-Algarve, Portugal, October 2012, pp. 573--580.

\bibitem{zhu2019improving}
Y.~Zhu, K.~Sapra, F.~A. Reda, K.~J. Shih, S.~Newsam, A.~Tao, and B.~Catanzaro,
  ``Improving semantic segmentation via video propagation and label
  relaxation,'' in \emph{Proc.~IEEE Conference on Computer Vision and Pattern
  Recognition}, Long Beach, CA, June 2019, pp. 8856--8865.

\bibitem{reda2018sdc}
F.~A. Reda, G.~Liu, K.~J. Shih, R.~Kirby, J.~Barker, D.~Tarjan, A.~Tao, and
  B.~Catanzaro, ``{SDC-N}et: Video prediction using spatially-displaced
  convolution,'' in \emph{Proc.~IEEE European Conference on Computer Vision},
  Munich, Germany, September 2018, pp. 718--733.

\bibitem{andreas2020kitti_semantic}
A.~Geiger, P.~Lenz, C.~Stiller, and R.~Urtasun, ``The {KITTI} semantic
  segmentation benchmark,'' {A}vailable from
  \url{http://www.cvlibs.net/datasets/kitti/eval_semseg.php?benchmark=semantics2015},
  2018.

\bibitem{Alhaija2018IJCV}
H.~Alhaija, S.~Mustikovela, L.~Mescheder, A.~Geiger, and C.~Rother, ``Augmented
  reality meets computer vision: Efficient data generation for urban driving
  scenes,'' \emph{International Journal of Computer Vision}, March 2018.

\bibitem{Geiger2012CVPR}
A.~Geiger, P.~Lenz, and R.~Urtasun, ``Are we ready for autonomous driving? {The
  KITTI} vision benchmark suite,'' in \emph{Proc.~IEEE Conference on Computer
  Vision and Pattern Recognition}, Providence, RI, June 2012, pp. 3354--3361.

\bibitem{romera2017erfnet}
E.~Romera, J.~M. Alvarez, L.~M. Bergasa, and R.~Arroyo, ``{ERFN}et: Efficient
  residual factorized convnet for real-time semantic segmentation,'' \emph{IEEE
  Transactions on Intelligent Transportation Systems}, vol.~19, no.~1, pp.
  263--272, Jan 2018.

\bibitem{Cordts2016Cityscapes}
M.~Cordts, M.~Omran, S.~Ramos, T.~Rehfeld, M.~Enzweiler, R.~Benenson,
  U.~Franke, S.~Roth, and B.~Schiele, ``The cityscapes dataset for semantic
  urban scene understanding,'' in \emph{Proc.~IEEE Conference on Computer
  Vision and Pattern Recognition}, Las Vegas, NV, June 2016, pp. 3213--3223.

\bibitem{stradvision_2021}
StradVision, ``https://stradvision.com/,'' 2021.

\bibitem{jetson}
{NVIDIA Jetson Modules},
  ``https://developer.nvidia.com/embedded/jetson-modules,'' 2021.

\bibitem{wang2019unified}
K.~Wang, Y.~Lin, L.~Wang, L.~Han, M.~Hua, X.~Wang, S.~Lian, and B.~Huang, ``A
  unified framework for mutual improvement of {SLAM} and semantic
  segmentation,'' in \emph{Proc.~IEEE International Conference on Robotics and
  Automation}, Montreal, QC, Canada, May 2019, pp. 5224--5230.

\end{thebibliography}

\vspace{20pc}

\begin{IEEEbiography}
[{\includegraphics[width=1in,height=1.25in,clip,keepaspectratio]{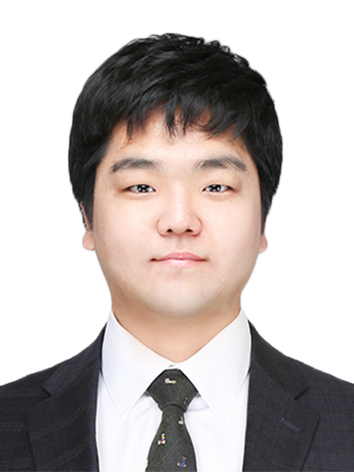}}]
{Jinkyu Lee}
was born in Chungju, Korea in 1993. He received B.S. degree in Computer Science from Handong Global University, Pohang, Korea, in 2019. He is currently pursuing M.S. degree in the Department of Information and Communication Engineering at the Handong Global University.

From 2016 to now, he is a research assistant with the computer graphics and vision laboratory. His current research interests are computer vision, simultaneous localization and mapping system, advanced driver assistance system, and autonomous driving.
\end{IEEEbiography}

 {\vskip -2\baselineskip plus -1fil}

\begin{IEEEbiography}
[{\includegraphics[width=1in,height=1.25in,clip,keepaspectratio]{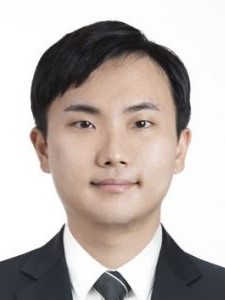}}]
{Muhyun Back}
was born in Gwangju, Korea in 1994. He received B.S. degree in Computer Science from Handong Global University, Pohang, Korea, in 2019. He is currently pursuing M.S. degree in the Department of Information and Communication Engineering at the Handong Global University.

From 2016 to now, he is a research assistant with the computer graphics and vision laboratory. His current research interests are computer vision, advanced driver assistance system, simultaneous localization and mapping system, and 3D reconstruction.
\end{IEEEbiography}

 {\vskip -2\baselineskip plus -1fil}

\begin{IEEEbiography}
[{\includegraphics[width=1in,height=1.25in,clip,keepaspectratio]{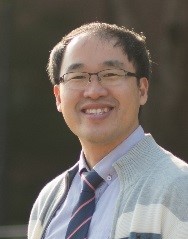}}]
{Sung Soo Hwang}
was born in Busan, South Korea in 1983. He received B.S. degree in Electrical Engineering and Computer Science from Handong Global Unveristy, Pohang, South Korea in 2008, and M.S and Ph. D degrees in Korea Advanced Institute of Science and Technology, Daejeon, South Korea in 2010 and 2015, respectively. 

He is an Associate Professor with School of Computer Science and Electrical Engineering, Handong Global University, Pohang, South Korea. His research interests include simultaneous localization and mapping system, autonomous driving, and image-based 3D modeling.
\end{IEEEbiography}    

 {\vskip -2\baselineskip plus -1fil}

\begin{IEEEbiography}[{\includegraphics[trim=0.08in 0.2in 0.08in 0in,height=1.25in,clip,keepaspectratio]{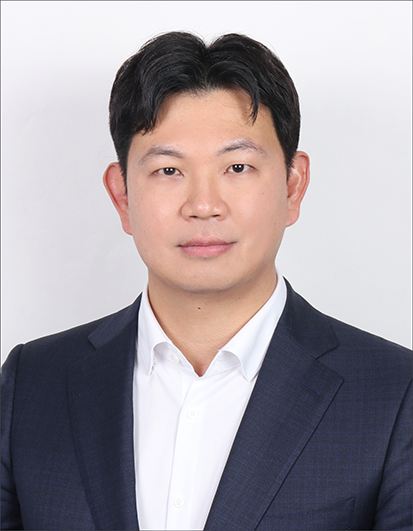}}]{Il Yong Chun}
(S'14--M'16) is a tenure-track Assistant Professor of EEE \& AI at Sungkyunkwan University (SKKU), Swuon, South Korea.
He received B.Eng. degree from Korea University in 2009,
and the Ph.D. degree from Purdue University in 2015,
both in electrical \& computer engineering.
During his Ph.D., he worked with Intel Labs, Samsung Advanced Institute of Technology, and Neuroscience Research Institute,
as a Research Intern or a Visiting Lecturer.
Prior to joining SKKU, he was a Postdoctoral Research Associate in Mathematics, Purdue University,
a Research Fellow in EECS, the University of Michigan,
and an Assistant Professor of ECE at the University of Hawai'i, M\=anoa,
from 2015 to 2016, from 2016 to 2019, from 2019 to 2021, respectively. 
His research interests include artificial intelligence \& machine learning, optimization, and compressed sensing, 
applied to applications in computational imaging, image processing, and computer vision. 
\end{IEEEbiography}

\end{document}